\documentclass{article}
\usepackage{arxiv} 

\usepackage[utf8]{inputenc} 
\usepackage[T1]{fontenc}    
\usepackage[numbers]{natbib}
\usepackage{lipsum}         
\usepackage{array,textcomp,url,comment} 
\usepackage{graphicx,subcaption}

\usepackage{etoolbox}
\usepackage{placeins}
\usepackage[table,xcdraw,dvipsnames]{xcolor}
\usepackage{pifont}
\newcommand{\cmark}{\ding{52}} 
\newcommand{\xmark}{\ding{53}} 
\usepackage{rotating}

\usepackage{amsfonts,amssymb,amsmath,amsthm,amsopn}	

\usepackage[T1,small,euler-digits]{eulervm}

\usepackage[linesnumbered,boxed,ruled,vlined]{algorithm2e} 
\definecolor{myblue2}{RGB}{70,131,180}

\SetCommentSty{mycommfont}
\SetKwInput{KwIn}{Input}
\SetKwInput{KwOut}{Output}

\usepackage{hhline,booktabs,colortbl,multirow,tabularx,threeparttable} 
\usepackage{enumitem, multicol}
\usepackage{xstring}
\usepackage{nicefrac}

\usepackage{enumerate}
\usepackage{footnote}
\usepackage{float}
\usepackage{hyperref}
\usepackage{prettyref}
\usepackage{setspace}
\usepackage{titlesec}

\definecolor{myblue}{RGB}{34,31,217}
\definecolor{mycyan}{gray}{.7}
\definecolor{Gray}{gray}{0.9}
\definecolor{headerblue}{RGB}{0,85,150}    
\definecolor{natureblue}{RGB}{11,61,145}   
\definecolor{titlegray}{RGB}{80,80,80}     
\definecolor{titlebox}{RGB}{245,248,252}   
\definecolor{myg}{gray}{0.95}
\definecolor{mylblue}{RGB}{210, 235, 255}

\newlength{\abstractpad}
\hypersetup{
  colorlinks=true,
  linkcolor=natureblue,
  citecolor=natureblue,
  urlcolor=natureblue,
  filecolor=natureblue
}

\title{A Firefly Algorithm for Mixed-Variable Optimization Based on Hybrid Distance Modeling}

\date{}

\usepackage{authblk}

\newbox{\orcid}\sbox{\orcid}{\includegraphics[scale=0.06]{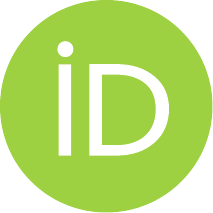}} 
\author[1]{%
	{Ousmane Tom Bechir}%
}
\author[1,2,$\ddagger$]{%
    \href{https://orcid.org/0000-0003-2623-5206}{\usebox{\orcid}}~{Adán José-García}%
}
\author[1,2]{%
	\\ \href{https://orcid.org/0000-0002-2551-6586}{\usebox{\orcid}}~{Zaineb Chelly Dagdia}%
}
\author[2,3]{%
	\href{https://orcid.org/0000-0003-3083-2441}{\usebox{\orcid}}~{Vincent Sobanski}%
}
\author[1]{%
    \href{https://orcid.org/0000-0002-6590-7215}{\usebox{\orcid}}~{Clarisse Dhaenens}%
}

\affil[1]{
Univ. Lille, CNRS, Inria, Centrale Lille, UMR  9189 CRIStAL, F-59000 Lille, France
}
\affil[2]{Univ. Lille, Inserm, CHU Lille, U1286 – INFINITE, Lille, France}
\affil[3]{
Institut Universitaire de France (IUF), Paris, France
}
\affil[$\ddagger$]{
Corresponding author: \texttt{adan.josegarcia@univ-lille.fr}
}

\begin{document}

\maketitle
\begin{abstract}

Several real-world optimization problems involve mixed-variable search spaces, where continuous, ordinal, and categorical decision variables coexist. However, most population-based metaheuristic algorithms are designed for either continuous or discrete optimization problems and do not naturally handle heterogeneous variable types. In this paper, we propose an adaptation of the Firefly Algorithm for mixed-variable optimization problems (FAmv). The proposed method relies on a modified distance-based attractiveness mechanism that integrates continuous and discrete components within a unified formulation. This mixed-distance approach enables a more appropriate modeling of heterogeneous search spaces while maintaining a balance between exploration and exploitation. 
The proposed method is evaluated on the CEC mixed-variable benchmark, which includes unimodal, multimodal, and composition functions. The results show that FAmv achieves competitive, and often superior, performance compared with state-of-the-art mixed-variable optimization algorithms. In addition, experiments on engineering design problems further highlight the robustness and practical applicability of the proposed approach.
These results indicate that incorporating appropriate distance formulations into the Firefly Algorithm provides an effective strategy for solving complex mixed-variable optimization problems.

\end{abstract}

\section{Introduction}

Optimization plays a key role across many domains, including energy systems, logistics, healthcare, and finance~\cite{boyd2004convex}. Real-world optimization problems often involve different types of decision variables, including continuous, discrete, and categorical variables. Such problems, commonly referred to as mixed-variable optimization problems (MVOPs), introduce additional challenges for modeling and computation~\cite{talbi2024metaheuristics}. While classical mathematical programming can solve some structured problems, many real-world cases are nonconvex, large-scale, or too complex to solve directly~\cite{burer2012non}. In such cases, metaheuristic algorithms offer a flexible and robust alternative. Among them, nature-inspired metaheuristics such as the Firefly Algorithm (FA)~\cite{yang2009firefly} have demonstrated good performance across a wide range of applications, including engineering design optimization, energy management, and parameter tuning~\cite{fister2013comprehensive}.

However, despite their success across many application domains, most metaheuristics were originally developed for either continuous or discrete problems (e.g., PSO, DE, GA)~\cite{pso, DE, ga}, making their direct application to mixed-variable optimization problems nontrivial. In particular, FA~\cite{yang2009firefly} was originally designed for continuous optimization, relying on a distance-based attractiveness mechanism defined using Euclidean distance, which does not naturally extend to heterogeneous variable types. As a result, its standard formulation is not well-suited to mixed-variable problems and requires more appropriate representations and interaction mechanisms.

Relaxation and discretization strategies are frequently used to adapt optimization algorithms for mixed-variable or discrete optimization problems. For instance, Gandomi et al.~\cite{classicalFAForMVOP} applied the classical FA to mixed-variable optimization using a relaxation approach, while in~\cite{DFAforTSP}, a discretized FA variant was proposed for the travelling salesman problem (TSP). Similarly, algorithms originally developed for discrete optimization are often extended to continuous or mixed-variable settings via encoding or discretization. A representative example is the genetic algorithm (GA)~\cite{ga}, which typically uses binary representations and can be applied to continuous domains via appropriate encoding schemes. These approaches, however, exhibit several limitations. In relaxation-based methods, discrete variables are embedded in a continuous space and subsequently mapped back using rounding or integer operators, which may lead to precision loss and suboptimal solutions. Similarly, discretization strategies may fail to adequately capture the structure of categorical variables. In the case of FA, existing adaptations remain limited to specific variable types (e.g., integer variables) and do not naturally extend to fully heterogeneous search spaces, including categorical variables. To the best of our knowledge, a general and unified adaptation of FA for mixed-variable optimization remains limited in the literature.

In this context, we propose an adaptation of FA for MVOPs based on type-aware movement mechanisms. Specifically, we extend the distance metric used in the attractiveness computation to combine continuous and discrete components. In addition, the random movement phase is modified to handle categorical variables, ensuring that candidate solutions remain valid within their respective domains. Furthermore, a parameter adaptation strategy is introduced to regulate the exploration–exploitation trade-off during the search process. These modifications enable efficient exploration of mixed-variable search spaces while preserving solution feasibility and precision.

The main contributions of this paper are as follows. First, we propose a type-aware adaptation of the FA for mixed-variable optimization. Second, we introduce mixed-distance formulations, including Euclidean–Hamming and Gower-based measures, to model interactions between solutions in heterogeneous search spaces. Third, we incorporate a parameter adaptation strategy to regulate the exploration-exploitation trade-off during the optimization process. Finally, we provide an experimental evaluation on synthetic and real-world mixed-variable optimization problems, including an ablation study that investigates the contribution of each component.



\section{Related Work} \label{sec:related_work}

\subsection{Mixed-Variable Optimization}

A mixed-variable optimization problem (MVOP) involves heterogeneous decision variables within a single optimization framework. The goal is to find a candidate solution that minimizes the objective function. We can formulate the problem as follows: 
\[
\min_{x \in \Omega} f(x)
\]
Where $\Omega$ is the search space, and $f$ is the objective function. When the search space $\Omega$ is heterogeneous, the decision vector can be expressed as:
\[ \label{enc:pos}
X = (x^{(c)}, x^{(d)})
\]
where $x^c \in \mathbb{R}^{n_c}$ represents the continuous component and $x^d$ denotes the discrete component, which may include integer, binary, or categorical variables. 

\subsection{Metaheuristic Approaches for Mixed-Variable Optimization}

Mixed-variable optimization problems (MVOPs) include heterogeneous decision variables, including continuous, integer, and categorical types, which makes the direct application of classical metaheuristics challenging. To address this, several extensions of evolutionary and swarm-based algorithms have been proposed to operate in mixed-variable search spaces.

A straightforward approach consists of using a unified encoding scheme, typically based on binary representations, to handle all variable types within a single framework~\cite{ga}. This type of representation allows the direct application of standard operators such as crossover and mutation. These encodings, however, often introduce discretization bias for continuous variables and increase the dimensionality of the search space, potentially leading to reduced precision and higher computational cost.

More advanced approaches rely on hybrid representations and type-specific operators. For instance, extensions of Differential Evolution (DE)~\cite{DE} to mixed-variable problems (DEmv) use continuous mutation and crossover for real-valued variables while applying specialized operators for discrete and categorical components~\cite{DEmv}. Similarly, Particle Swarm Optimization (PSO)~\cite{pso} has been adapted to mixed-variable settings (PSOmv) by combining classical velocity updates for continuous variables with probabilistic update rules for discrete components~\cite{psomv}. Ant Colony Optimization (ACO) has also been extended to heterogeneous spaces by designing mechanisms that simultaneously explore continuous and discrete variables~\cite{ACO_mv}. In addition, model-based approaches such as estimation-of-distribution algorithms (EDA) have been proposed for MVOPs by explicitly modeling the joint distribution of heterogeneous variables~\cite{EDA_mv}. These methods can capture dependencies between variables but often require complex modeling assumptions and increased computational effort.

Although significant progress has been made, most existing approaches primarily adapt representations or develop type-specific operators, while the interaction mechanisms between candidate solutions remain largely inherited from their original formulations. Specifically, the relationships between continuous and discrete variables are frequently poorly modeled, which can restrict the effectiveness of these methods in exploring heterogeneous search spaces.

\subsection{The Firefly Algorithm}

The Firefly Algorithm (FA) is a population-based metaheuristic inspired by the flashing behavior of fireflies, where each firefly represents a candidate solution in the search space~\cite{yang2009firefly}. The algorithm assumes that all fireflies are unisex and attracted to one another. The attractiveness of the firefly is proportional to its brightness, and it decreases with distance. The brightness of a firefly is determined by the landscape of the objective function. 
The main steps of the algorithm are described below, while its pseudo-code is provided in Algorithm~\ref{alg:firefly} in Appendix~\ref{app:FA_algorithm}.

Let us consider a population of $N$ fireflies $\{X_i\}_{i=1}^N$. The attractiveness between two fireflies $X_i$ and $X_j$ ($i \neq j$) in a continuous space is defined as in~\cite{yang2009firefly}:

\begin{equation}\label{eq:attractiveness}
    \beta(r_{i,j}) = \beta_0 e^{-\gamma r_{i,j}^2}
\end{equation}
where $\beta_0$ is the attractiveness at $r=0$, $\gamma$ is the light absorption coefficient, and $r_{i,j}$ denotes the distance between two fireflies, typically defined as the Euclidean distance:

\begin{equation} \label{eq:euclidean_distance}
    r_{i,j} = \sqrt{\sum_{k=1}^D \left(X_j^{(k)} - X_i^{(k)}\right)^2}
\end{equation}
where $D$ denotes the dimension of the problem. 

The movement of a firefly $X_i$ attracted to a brighter firefly $X_j$ is defined as in~\cite{yang2009firefly}:

\begin{equation} \label{eq:update_position}
    X_i(t+1) = X_i(t) + \beta(r_{i,j})\bigl(X_j(t) - X_i(t)\bigr) + \alpha\bigl(\mathrm{rand}-\tfrac{1}{2}\bigr)
\end{equation}

In Eq.~\eqref{eq:update_position}, the second term represents the attraction toward brighter fireflies, while the third term introduces a random perturbation controlled by the parameter $\alpha$, where $\mathrm{rand}$ is a random number uniformly distributed in $[0,1]$.

\subsection{Extensions of the Firefly Algorithm}

Several extensions of FA have been proposed to improve its applicability and performance. A first line of work focuses on adapting FA to discrete or combinatorial optimization problems. For instance, discrete variants replace the Euclidean distance with measures such as the Hamming distance and introduce problem-specific movement operators to preserve feasibility~\cite{DFA_PWREnhancement, Durkota2011}. These approaches enable FA to operate in discrete spaces but are typically limited to specific variable types. A second line of research investigates adaptive and memetic strategies. For example, Fister et al.~\cite{fister2013memetic} proposed a self-adaptive FA in which control parameters are dynamically adjusted during the optimization process. These methods improve robustness and help balance exploration and exploitation. A third direction focuses on improving efficiency and maintaining population diversity. Methods such as the Neighborhood Enhanced FA (NEFA)~\cite{NEFA} reduce computational cost by restricting interactions to subsets of individuals, whereas clustering-based approaches, such as INEFA~\cite{INEFA}, promote diversity by structuring the population into groups.

Despite recent advances, current FA variants exhibit limitations when addressing mixed-variable optimization problems. Most existing methods adapt operators to specific variable types or employ discretization strategies, yet they do not redefine the underlying interaction mechanism. In particular, the conventional distance-based attractiveness formulation is not inherently compatible with heterogeneous search spaces that include continuous, discrete, and categorical variables. This challenge motivates the need for a unified, type-aware interaction mechanism for FA in mixed-variable search spaces.

\section{Proposed Approach} \label{sec:proposed_approach}

We propose a Firefly Algorithm for mixed-variable optimization problems, named \textbf{FAmv}. In this approach, a solution is represented as a structured vector comprising continuous and discrete components, with heterogeneous dimensions handled using type-specific operators. Unlike existing approaches that adapt the representation to fit classical operators~\cite{DEmixed, psomv}, FAmv preserves the original firefly dynamics and introduces dedicated update rules for each variable type. This decoupled design allows precise control of both continuous and discrete movements while maintaining structural feasibility. More specifically, the proposed approach introduces the following components: (i) a mixed-variable distance model to guide interactions between solutions, (ii) a discrete movement mechanism based on probabilistic attraction and exploration, and (iii) a unified interaction framework integrating continuous and discrete components within a single search process. The proposed FAmv is outlined in Algorithm~\ref{alg:FAmv}, and each main component is detailed in the following sections.

\begin{algorithm}[htb!]
\DontPrintSemicolon
\setstretch{1.0}

Randomly initialize the population of $N$ fireflies, $X_1, X_2, \ldots, X_N$\;
Compute the fitness value of each firefly using $f$\;
Initialize the global best positions and fitness values\;
$t \leftarrow 1$\;

\While{$t \leq MAX\_ITER$} {
    Update parameters $\alpha$ and $\gamma$ if an adaptive strategy is used\;
    \For{$i \leftarrow 1$ \KwTo $N$}{
        updated $\leftarrow$ \texttt{false}\;
        \For{$j \leftarrow 1$ \KwTo $N$}{
            \If{$f(X_j) < f(X_i)$}{
                Compute distance $R_{i,j}$ using Eq.~\eqref{eq:dist_E_H} or Eq.~\eqref{Eq:gower_dist}\;
                Extract the continuous components $x_i^{(c)}$ and $x_j^{(c)}$ from $X_i$ and $X_j$\;
                Extract the discrete components $x_i^{(d)}$ and $x_j^{(d)}$ from $X_i$ and $X_j$\;
                Compute attractiveness $\beta(r_{i,j})$ for the continuous position using Eq.~\eqref{eq:attractiveness}\;
                Compute attractiveness probability $\beta_{\mathrm{prob}}(r_{i,j})$ for the discrete position using Eq.~\eqref{Eq:attrac_prob}\;
                $x_i^{(c)} \leftarrow x_i^{(c)} + \beta(r_{i,j})(x_j^{(c)} - x_i^{(c)}) + \alpha\left(\mathrm{rand} - \nicefrac{1}{2}\right)$\;
                Move $x_i^{(d)}$ toward $x_j^{(d)}$ using the $\beta$-step and $\alpha$-step rules\;
                $X_i \leftarrow \left(x_i^{(c)}, x_i^{(d)}\right)$\tcp*[r]{combine continuous and discrete components}
                Compute the fitness value of the new $X_i$\;
                Update the global best positions and fitness values\;
                updated $\leftarrow$ \texttt{true}\;
            }
        }
        \If{updated $=$ \texttt{false}}{
            $x_i^{(c)} \leftarrow x_i^{(c)} + \alpha\left(\mathrm{rand} - \nicefrac{1}{2}\right)$\tcp*[r]{random walk in the continuous space}
            Add a random walk to $x_i^{(d)}$ using the $\alpha$-step rules\;
            $X_i \leftarrow \left(x_i^{(c)}, x_i^{(d)}\right)$\tcp*[r]{combine continuous and discrete components}
            Compute the fitness value of the new $X_i$\;
            Update the personal and global best positions and fitness values\;
        }
    }
    $t \leftarrow t + 1$ \tcp*[r]{stopping criterion can also be defined based on function evaluations (FE)}
}

\caption{\label{alg:FAmv}FAmv: Firefly algorithm for mixed variables}
\end{algorithm}

\subsection{Attractiveness Modeling}

The original FA~\cite{yang2009firefly, DFAforTSP} considers the Euclidean distance in Eq.~\eqref{eq:attractiveness} as a component to compute the attractiveness between fireflies, which is well-suited to numeric search spaces. However, in discrete-dimensional search spaces, the original formulation can be misleading, degrading the algorithm's performance when exploring such heterogeneous, complex spaces. In particular, Euclidean distance does not account for the structural differences induced by categorical or discrete variables, which may lead to inappropriate similarity measures between candidate solutions. In this direction, we present two new strategies to better model the attraction between fireflies in heterogeneous search spaces.

\paragraph{Attractiveness based on Hamming distance} We propose a unified mixed distance that operates across both subspaces (numeric and discrete spaces). This distance is defined as follows:

\begin{equation} \label{eq:dist_E_H}
    r_{i,j} = \frac{1}{D}(d_E(x_i^{(c)}, x_j^{(c)}) + d_H(x_i^{(d)}, x_j^{(d)}))
\end{equation}
where $d_E$ is the Euclidean distance over continuous, numeric variables; $d_H$ is the Hamming distance over discrete, categorical variables; and $D$ is the total number of dimensions. The normalization by $D$ is introduced to ensure that the resulting distance remains bounded and comparable across problems of different dimensionalities. However, this formulation does not explicitly balance the relative contributions of continuous and discrete components, leading to different interaction behaviors depending on the scale of the variables.

\paragraph{Attractiveness based on Gower distance} We consider the Gower distance, a well-established mixed-variable metric widely used in cluster analysis, to measure similarity between data points with heterogeneous variable types. It is particularly well-suited to mixed-variable optimization problems, as it normalizes each variable’s contribution independently before aggregation. The Gower distance between two fireflies $X_i$ and $X_j$ is defined as follows~\cite{gower1971general}:

\begin{equation}\label{Eq:gower_dist}
    r_{i,j} = \frac{1}{D}\sum_{k=1}^D \delta_{i,j}^k,
\end{equation}
where $\delta_{i,j}^k$ represents the normalized contribution of the $k$-th variable to the overall distance, and is defined according to the data type of the $k$-th variable as:

\[
\delta_{i,j}^k =
\begin{cases}
\dfrac{|x_{i}^{(k)} - x_{j}^{(k)}|}{\mathrm{range}_k} & \text{if } k \text{ is continuous} , \\[8pt]
1 & \text{if } k \text{ is discrete (categorical or integer)} , \\[4pt]
0 & \text{otherwise, if } x_i^{(k)} = x_j^{(k)} , 
\end{cases}
\]
where $\mathrm{range}_k$ denotes the range of the $k$-th variable. This formulation ensures that each variable contributes to the overall distance in proportion to its scale or type, leading to a more balanced interaction among heterogeneous components.

Beyond their role in computing attractiveness, these two proposed distance functions directly influence the interaction dynamics of our proposed FA framework. Since FA is inherently distance-based, the proposed mixed-distance formulations implicitly allow information exchange across continuous and discrete components during attraction. Thus, by aggregating contributions from both subspaces within a unified metric, the movement intensity reflects variations occurring in either part of the solution. This unified formulation enables the attraction mechanism to jointly account for heterogeneous components when guiding the search process. We therefore hypothesize that this mechanism better reflects the joint influence of continuous and discrete variables, leading to more coherent search trajectories in heterogeneous landscapes.

\subsection{Mixed-Variable Movement Mechanism} 
\label{subsec:famv_movement}

We propose movement mechanisms for fireflies that are particularly suited to mixed-variable optimization problems. These mechanisms are introduced to ensure safe and coherent handling of heterogeneous variables. We decouple the movement mechanisms into continuous and discrete updates, allowing each variable type to be updated according to its intrinsic structure while maintaining a unified interaction process.

First, for continuous variables, the update is performed identically to that in the original FA: attractiveness using Eq.~\eqref{eq:attractiveness} based on Euclidean distance, followed by the standard update rule Eq.~\eqref{eq:update_position}. Then, for discrete variables, we propose an update mechanism (inspired by the strategy introduced in~\cite{DFA_PWREnhancement}) that comprises a two-phase process: a guided attraction phase ($\beta$-step) and a random exploration phase ($\alpha$-step), which respectively promote exploitation and exploration in the discrete search space.

\paragraph{$\beta$-Step} Its objective is to move the discrete part of a firefly $x_i^{(d)}$ toward a brighter firefly $x_j^{(d)}$. The attraction strength ($r_{i,j}$) is computed using one of the proposed distances, Eq.~\eqref{eq:dist_E_H} or Eq.~\eqref{Eq:gower_dist}. Therefore, the attractiveness coefficient is defined as:

\begin{equation} \label{Eq:attrac_prob}
    \beta = \exp(-\gamma r_{i,j}^2)
\end{equation}

This value is interpreted as the probability of exchange between the discrete components of the two interacting fireflies, thereby extending the FA attraction mechanism to discrete variables in a probabilistic manner. For each discrete dimension $k$, such that $x_{i,k}^{(d)} \neq x_{j,k}^{(d)}$, the value of $x_{i,k}^{(d)}$ may be replaced by $x_{j,k}^{(d)}$ with probability $\beta$. This probabilistic replacement can be interpreted as a distance-guided discrete movement operator, where the intensity of the modification is controlled by the relative proximity between solutions. When two fireflies are close in the mixed search space, the distance $r_{i,j}$ tends to be small, leading to a large value of $\beta$; therefore, more discrete components are likely to be exchanged. On the contrary, when two fireflies are far apart, $r_{i,j}$ increases, $\beta$ decreases, and, therefore, fewer discrete components are modified. Finally, discrete components that are already identical between $x_i$ and $x_j$ remain unchanged. Only differing components are subject to probabilistic exchange.

\paragraph{$\alpha$-Step} The objective of the $\alpha$-step is to introduce a random movement in order to promote exploration in the search space. To preserve the FA's original dynamics, we reuse the $\alpha$-step mechanism from the original work. However, in our mixed-variable setting, we divide the discrete $\alpha$-step into two cases depending on the nature of the variable. This distinction is introduced to ensure that the exploration mechanism remains consistent with the variable domain's structure. First, for integer variables defined over compact intervals (i.e., consecutive integer values), we directly adopt the perturbation mechanism used in the discrete firefly formulation. The update is defined as:

\begin{equation} \label{Eq:alpha-step1}
    x_{i,k}^{(d)} = \mathrm{INT}(x_{i,k}^{(d)} + \alpha \cdot \varepsilon)~,
\end{equation}
where $\varepsilon$ is a random value sampled from a given interval. This formulation preserves local exploration by generating small perturbations around the current solution. Secondly, for discrete variables not belonging to compact integer intervals (e.g., categorical variables or integer domains with gaps), we redefine the $\alpha$-step as a probabilistic random replacement mechanism, using the following function:

\begin{equation} \label{Eq:prob_alpha_adapt}
    p_{\alpha} = \frac{1}{1 + \exp\left(-k \left(\alpha - \frac{\alpha_\mathrm{init}}{2}\right)\right)}~,
\end{equation}
where $k$ is a control hyperparameter. This formulation maps the exploration parameter $\alpha$ to a probability of replacement, allowing a smooth transition between low and high exploration states. When a constant $\alpha$ is used, the transformation becomes:

\begin{equation}  \label{Eq:prob_alpha_const}
p_{\alpha} = \frac{1}{1 + \exp\left(-k \frac{\alpha}{2}\right)}.
\end{equation}

The parameter $k$ controls the transition behavior between exploration and exploitation. Larger values of $k$ yield a sharper transition, whereas smaller values yield a smoother transition. Figure~\ref{fig:p_alpha} shows the influence of $k$ on the transition from exploration to exploitation. Using this probability, each non-compact discrete variable is randomly replaced, with probability $p_{\alpha}$, by another value sampled from its corresponding distribution, thus enabling global exploration in discrete domains where local perturbations are not meaningful.

\subsection{Parameter Adaptation}
\label{subsec:famv_adaptation}

FA explores the search space using the parameters $\gamma$ and $\alpha$, for exploration and exploitation, respectively. On one hand, the parameter $\gamma$ controls the decay of attractiveness. A small value leads to strong exploitation, meaning that fireflies are strongly influenced by others even at relatively large distances. Conversely, for large $\gamma$, the influence decreases rapidly with distance, thereby reducing the intensity of exploitation. On the other hand, the $\alpha$ parameter controls the perturbation level. Large $\alpha$ values introduce stronger noise in both continuous and discrete updates, leading to greater exploration. On the contrary, smaller $\alpha$ values reduce the magnitude (or probability) of random movements, thus limiting exploration. In the context of mixed-variable optimization, these parameters jointly regulate the interaction between continuous and discrete components, influencing how the search process balances global exploration and local refinement across heterogeneous spaces.

In this work, we investigate adaptive parameter strategies for both $\alpha$ and $\gamma$. Their adaptation is based on information about the optimization progress. This choice is motivated by the need to dynamically balance exploration and exploitation in mixed-variable search spaces, where different variable types may require different search behaviors during the optimization process. The parameters $\alpha$ and $\gamma$ are updated at each iteration as follows:

\begin{equation} \label{Eq:adapt_alpha}
\alpha = \max\left(0.01, \alpha_{\mathrm{init}} \cdot (1 - \text{progress}) \right)~,
\end{equation}

and

\begin{equation} \label{Eq:adapt_gamma}
\gamma = \max\left(0.01, \gamma_{\mathrm{init}} \cdot (1 - \text{progress}) \right)~,
\end{equation}
where $\mathrm{progress} = \nicefrac{\mathrm{FE}}{\mathrm{FE}_{\max}}$ is defined as the ratio between the number of consumed function evaluations (FE) and the maximum evaluation budget ($\mathrm{FE}_{\max}$). The lower bound ($0.01$) is introduced to prevent the exploration level from becoming zero, which could lead to premature convergence. This schedule ensures that a minimum level of stochasticity is maintained throughout the search process.

Both parameters $\alpha$ and $\gamma$ are therefore initialized to typical values and then progressively decreased during optimization. The underlying idea is to encourage stronger exploration at the beginning of the search, when the available budget is mostly unused, and to gradually increase exploitation as the algorithm approaches the maximum evaluation limit. In particular, decreasing $\alpha$ reduces random perturbations, while decreasing $\gamma$ increases the locality of interactions, thereby jointly enabling a transition from global exploration to local refinement.



\section{Experimental Setup} \label{sec:experimental_setup}

In this section, we present the experimental settings used to evaluate the effectiveness of our proposed FAmv compared with three baseline algorithms.

\subsection{Baseline Algorithms} \label{subsec:setup_baselines}

We considered three state-of-the-art MVO algorithms in our experiment: DEmv~\cite{DEmixed}, based on differential evolution; GA~\cite{mitchell1998introduction}, based on a binary-encoded genetic algorithm; and PSOmv~\cite{psomv}, from the swarm optimization family. These algorithms were implemented according to their original papers. 

Four variants of our proposed FAmv algorithm are evaluated in order to investigate the impact of certain design choices: FAmv$_\mathrm{H}$ is based on the mixed Euclidean-Hamming as defined in Eq.~\eqref{eq:dist_E_H}, and FAmv$_\mathrm{G}$ is based on the Gower distance as defined in Eq.~\eqref{Eq:gower_dist}. In these two versions, the control parameters $\alpha$ and $\gamma$ are fixed. Similarly, we also evaluate their respective adaptive versions, FAmv$^{*}_\mathrm{H}$ and FAmv$^{*}_\mathrm{G}$, indicating that $\alpha$ and $\gamma$ are adapted as detailed in Section~\ref{subsec:famv_movement}. These four configurations allow us to analyze the influence of the proposed attractiveness functions and the impact of parameter adaptation separately. A detailed justification of these design choices is provided after analyzing the impact of the mixed-variable movement mechanism in FA in Section~\ref{subsec:ablation_study}.

\paragraph{Parameter Settings} The FAmv population size was set to $25$ individuals across all variants, where the attractiveness at zero distance ($\beta_0$) was set to $1.5$. The versions FAmv$_\mathrm{H}$ and FAmv$_\mathrm{G}$ considered $\alpha = 1.5$ and $\gamma = 0.1$. In the adaptive versions, FAmv$^{*}_\mathrm{H}$ and FAmv$^{*}_\mathrm{G}$, the parameter $\alpha_{\mathrm{init}}$ starts at $2$, whereas $\gamma_{\mathrm{init}}$ starts at $0.05$.
For DEmv, PSOmv and GA, the population size was set to 100 individuals. In DEmv, the crossover rate $C$ and the differential weight $F$ are set to 0.2 as indicated in the original paper~\cite{DEmixed}. Similarly, for PSOmv, we followed the configuration settings specified in~\cite{psomv}. Finally, GA used the one-point crossover operator, the tournament selection mechanism, and an elitist strategy: the crossover and mutation probabilities were set to 0.9 and 0.01, respectively, and the tournament size was three~\cite{mitchell1998introduction}. All algorithms are compared under the same function-evaluation (FE) budget, providing a fair comparison of computational cost despite differences in population sizes.

\subsection{Benchmark Problems}

In our experiments, we used the CEC 2013 benchmark suite \cite{CEC2013benchmark}, a widely used benchmark for continuous optimization, comprising 28 mathematical functions. Additionally, we considered three engineering design problems with mixed-variable formulations of varying complexity: VESSEL~\cite{yang2010firefly_VESSEL}, BEAM~\cite{coello2000useBEAM}, and CSD~\cite{belegundu1982optimization_csd}.

\paragraph{CEC Synthetic Benchmark} This benchmark suite comprises 28 continuous optimization functions~\cite{CEC2013benchmark}. The suite includes five unimodal functions, fifteen basic multimodal functions, and eight composition functions. Originally, all variables in this benchmark are continuous. Therefore, to assess the algorithms' performance in a mixed-variable setting, we transformed the original formulation into a 50-dimensional mixed-variable search space, as in~\cite{DEmixed, psomv}. Among the 50 variables, the first 25 variables remain continuous and preserve exactly the same properties, ranges, and numerical characteristics as defined in the CEC specification. The remaining 25 variables are forced to take integer values; therefore, the MVO algorithms are restricted to exploring only integer values along those dimensions. This transformation follows previous works~\cite{DEmixed, psomv} and provides a controlled setting to analyze the behavior of algorithms in the presence of heterogeneous variable types. As this construction represents a synthetic mixed-variable scenario, we complement this evaluation with real mixed-variable engineering problems. The characteristics of these functions are presented in Table~\ref{tab:CEC2013}, whereas the mathematical definitions can be found in~\cite{CEC2013benchmark}.

\begin{table}[h!]
\caption{The 28 benchmark functions from CEC 2013 and their main characteristics. The presence or absence of a property is indicated by \cmark or \xmark, respectively. Continuous denotes functions defined over continuous domains; separable indicates that the objective can be decomposed into independent contributions per variable; symmetric refers to invariance under permutation or sign changes of variables.}
\centering
\small
\renewcommand{\arraystretch}{1.15}
\setlength{\tabcolsep}{0.1cm}
\begin{tabularx}{\textwidth}{@{} l c X c c c c @{}}
\toprule
\textbf{Category} & \textbf{Function ID} & \textbf{Function Name} & \textbf{Continuous} & \textbf{Separable} & \textbf{Symmetric} & \textbf{Optimum} \\
\midrule
\multirow{5}{*}{Unimodal}
  & \textbf{F1}  & Sphere Function                                   & \cmark & \cmark & \cmark & -1400 \\
  & \textbf{F2}  & Rotated High Conditioned Elliptic Function        & \cmark & \xmark & \cmark & -1300 \\
  & \textbf{F3}  & Rotated Bent Cigar Function                       & \cmark & \xmark & \xmark & -1200 \\
  & \textbf{F4}  & Rotated Discus Function                           & \cmark & \xmark & \xmark & -1100 \\
  & \textbf{F5}  & Different Powers Function                         & \cmark & \cmark & \cmark & -1000 \\[0.2em]
\midrule
\multirow{15}{*}{Multimodal}
  & \textbf{F6}  & Rotated Rosenbrocks Function                      & \cmark & \xmark & \xmark & -900 \\
  & \textbf{F7}  & Rotated Schaffers F7 Function                     & \cmark & \xmark & \xmark & -800 \\
  & \textbf{F8}  & Rotated Ackleys Function                          & \cmark & \xmark & \xmark & -700 \\
  & \textbf{F9}  & Rotated Weierstrass Function                      & \cmark & \xmark & \xmark & -600 \\
  & \textbf{F10} & Rotated Griewanks Function                        & \cmark & \xmark & \cmark & -500 \\
  & \textbf{F11} & Rastrigins Function                               & \cmark & \cmark & \xmark & -400 \\
  & \textbf{F12} & Rotated Rastrigins Function                       & \cmark & \xmark & \xmark & -300 \\
  & \textbf{F13} & Non-Continuous Rotated Rastrigins Function        & \xmark & \xmark & \xmark & -200 \\
  & \textbf{F14} & Schwefel's Function                               & \cmark & \xmark & \xmark & -100 \\
  & \textbf{F15} & Rotated Schwefel's Function                       & \cmark & \xmark & \xmark & 100 \\
  & \textbf{F16} & Rotated Katsuura Function                         & \cmark & \xmark & \xmark & 200 \\
  & \textbf{F17} & Lunacek-Bi-Rastrigin Function                     & \cmark & \cmark & \xmark & 300 \\
  & \textbf{F18} & Rotated Lunacek Bi-Rastrigin Function             & \cmark & \xmark & \xmark & 400 \\
  & \textbf{F19} & Expanded Griewanks plus Rosenbrocks Function      & \cmark & \xmark & \cmark & 500 \\
  & \textbf{F20} & Expanded Scaffers F6 Function                     & \cmark & \xmark & \xmark & 600 \\[0.2em]
\midrule
\multirow{8}{*}{Composition}
  & \textbf{F21} & Composition Function 1 (n=5, Rotated)             & \cmark & \xmark & \xmark & 700 \\
  & \textbf{F22} & Composition Function 2 (n=3, Unrotated)           & \cmark & \xmark & \xmark & 800 \\
  & \textbf{F23} & Composition Function 3 (n=3, Rotated)             & \cmark & \xmark & \xmark & 900 \\
  & \textbf{F24} & Composition Function 4 (n=3, Rotated)             & \cmark & \xmark & \xmark & 1000 \\
  & \textbf{F25} & Composition Function 5 (n=3, Rotated)             & \cmark & \xmark & \xmark & 1100 \\
  & \textbf{F26} & Composition Function 6 (n=5, Rotated)             & \cmark & \xmark & \xmark & 1200 \\
  & \textbf{F27} & Composition Function 7 (n=5, Rotated)             & \cmark & \xmark & \xmark & 1300 \\
  & \textbf{F28} & Composition Function 8 (n=5, Rotated)             & \cmark & \xmark & \xmark & 1400 \\
\bottomrule
\end{tabularx}
\label{tab:CEC2013}
\end{table}

\paragraph{Pressure Vessel design (VESSEL)} This is a four-dimensional constrained optimization problem~\cite{yang2010firefly_VESSEL}. Two variables correspond to discrete thickness parameters that must be multiples of 0.0625, while the remaining two variables are continuous. The decision vector is $x = (d_s, d_h, r, L)$, where $d_s$ and $d_h$ denote shell and head thicknesses (discrete), and $r$ and $L$ represent the inner radius and the length of the cylindrical section (continuous). The objective function minimizes the total fabrication cost and is defined as:
\[
f(x) = 0.6224\, r d_s L + 1.7781\, d_h r^2 + 3.1661\, d_s^2 L + 19.84\, d_s^2 r.
\]

The problem is subject to four inequality constraints related to stress limits, volume requirements, and geometric limitations; therefore, a penalty-based constraint-handling approach is adopted. When constraint violations occur, a large penalty term is added to the objective function. The complete mathematical formulation of the objective and constraints is summarized in Table~\ref{tab:vessel}.

\begin{table*}[th!]
\centering
\caption{Mathematical formulation of the Pressure Vessel design problem.}
\label{tab:vessel}

\begin{tabularx}{\textwidth}{l X c}
\toprule
\textbf{Variable} & \textbf{Description} & \textbf{Domain} \\
\midrule

$d_s$ & Shell thickness (discrete variable) & multiple of $0.0625$ \\
$d_h$ & Head thickness (discrete variable) & multiple of $0.0625$ \\
$r$   & Inner radius of the pressure vessel & $10 \le r \le 200$ \\
$L$   & Length of the cylindrical section & $10 \le L \le 200$ \\

\addlinespace
\midrule
\multicolumn{3}{l}{\textbf{Constraints}} \\
\midrule

$g_1(x)$ & $-d_s + 0.0193r \le 0$ &  \\
$g_2(x)$ & $-d_h + 0.00954r \le 0$ &  \\
$g_3(x)$ & $-\pi r^2 L - \frac{4}{3}\pi r^3 + 1296000 \le 0$ &  \\
$g_4(x)$ & $L - 240 \le 0$ &  \\

\bottomrule
\end{tabularx}

\end{table*}

\paragraph{Welded Beam Design (BEAM)} This is a four-dimensional constrained optimization problem involving only continuous variables~\cite{coello2000useBEAM}. The objective is to minimize the fabrication cost of a welded beam subject to stress, buckling, and deflection constraints. The objective function is defined as:
\[
f(x) = 1.10471 x_1^2 x_2 + 0.04811 x_3 x_4 (14 + x_2),
\]
where the variables represent: weld thickness ($x_1$), weld length ($x_2$), beam height ($x_3$), and beam thickness ($x_4$). Several nonlinear constraints are imposed, including shear stress limits, normal stress limits, geometric relations, buckling load constraints, and maximum deflection requirements. Constraint violations are handled using a penalty formulation that combines the violation magnitude with the number of violated constraints. The mathematical description is provided in Table~\ref{tab:beam}.

\begin{table*}[t]
\centering
\caption{Mathematical formulation of the Welded Beam design problem.}
\label{tab:beam}

\begin{tabularx}{\textwidth}{l X c}
\toprule
\textbf{Variable} & \textbf{Description} & \textbf{Domain} \\
\midrule

$x_1$ & Weld thickness & continuous \\
$x_2$ & Weld length & continuous \\
$x_3$ & Beam height & continuous \\
$x_4$ & Beam thickness & continuous \\

\addlinespace
\midrule
\multicolumn{3}{l}{\textbf{Constraints}} \\
\midrule

$g_1(x)$ & $\tau(x) - \tau_{\max} \le 0$ &  \\
$g_2(x)$ & $\sigma(x) - \sigma_{\max} \le 0$ &  \\
$g_3(x)$ & $x_1 - x_4 \le 0$ &  \\
$g_4(x)$ & $0.10471x_1^2 + 0.04811x_3x_4(14 + x_2) - 5 \le 0$ &  \\
$g_5(x)$ & $0.125 - x_1 \le 0$ &  \\
$g_6(x)$ & $\delta(x) - \delta_{\max} \le 0$ &  \\
$g_7(x)$ & $P - P_c(x) \le 0$ &  \\

\bottomrule
\end{tabularx}

\end{table*}

\paragraph{Coil Spring Design (CSD)} This is a three-dimensional constrained optimization problem involving two continuous variables and one discrete variable~\cite{belegundu1982optimization_csd}. The decision vector is defined as $x = (d, D, N)$, where $d$ is the wire diameter, $D$ is the mean coil diameter, and $N$ is the number of active coils. The objective function minimizes the weight of the spring and is defined as:
\[
f(x) = (N + 2) d^2 D.
\]

The problem includes multiple nonlinear inequality constraints on shear stress, surge frequency, minimum deflection, geometric ratios, and maximum allowable deflection. A penalty method is used to incorporate constraint violations into the objective value. The complete mathematical formulation is summarized in Table~\ref{tab:csd}.

\begin{table*}[t]
\centering
\caption{Mathematical formulation of the Coil Spring design (CSD) problem.}
\label{tab:csd}

\begin{tabularx}{\textwidth}{l X c}
\toprule
\textbf{Variable} & \textbf{Description} & \textbf{Domain} \\
\midrule

$d$ & Wire diameter & continuous, $d \ge 0.2$ \\
$D$ & Mean coil diameter & continuous, $D \le 3.0$ \\
$N$ & Number of active coils & discrete \\

\addlinespace
\midrule
\multicolumn{3}{l}{\textbf{Constraints}} \\
\midrule

$g_1(x)$ & $\dfrac{8 C_f P_{\max} D}{\pi d^3} - S \le 0$ &  \\
$g_2(x)$ & $\delta_{\max} + 1.05 (N + 2)d - L_{\text{free}} \le 0$ &  \\
$g_3(x)$ & $0.2 - d \le 0$ &  \\
$g_4(x)$ & $(d + D) - 3.0 \le 0$ &  \\
$g_5(x)$ & $3 - \dfrac{D}{d} \le 0$ &  \\
$g_6(x)$ & $\delta_{\max} - \delta_{pm} \le 0$ &  \\
$g_7(x)$ & $\delta_w - \delta_{\max} + \delta_{\text{load}} \le 0$ &  \\

\bottomrule
\end{tabularx}

\end{table*}

\subsection{Evaluation Protocol}

The results reported in Section~\ref{sec:results} are statistics computed from 30 independent executions for each MVO algorithm considered in our study. The performance of each algorithm is evaluated by the absolute error between the function's theoretical optimum and the best objective value achieved by the algorithm. For the synthetic benchmark functions in the CEC2013 suite, a computational budget of 100\,000 objective function evaluations (FEs) was set per execution, whereas a smaller budget of 10\,000 FEs was used for the engineering design problems due to their lower dimensionality. All algorithms were evaluated under the same FE budget per problem.

In our experiments, to provide a statistical interpretation of the significance of the comparison results, we use the following statistical measures. First, the non-parametric Kruskal-Wallis test \cite{kruskal1952use} is used to investigate the overall differences among the algorithms. Then, if significant differences are observed, the Holm correction \cite{holm1979simple} procedure is applied to control the family-wise error rate. Pairwise comparisons between algorithms were then performed using Dunn's post-hoc test \cite{dunn1964multiple}. All statistical tests are conducted at a significance level of $\alpha = 0.05$.

\section{Results} \label{sec:results}

This section discusses the results of a series of experiments conducted to investigate the performance of our mixed-variable optimization algorithm, FAmv. First, Section~\ref{subsec:results_benchmark_CEC2013} presents an overall evaluation of the performance and robustness of our algorithm and the baseline approaches when dealing with the CEC2013 benchmark. Then, the effectiveness of our FAmv variants in engineering design problems is studied in Section~\ref{subsec:results_engineering_problems}. Finally, Section~\ref{subsec:ablation_study} explores the impact of the main design components of FAmv through an ablation analysis.

\begin{figure} [h!]
    \centering
    \includegraphics[width=0.95\linewidth]{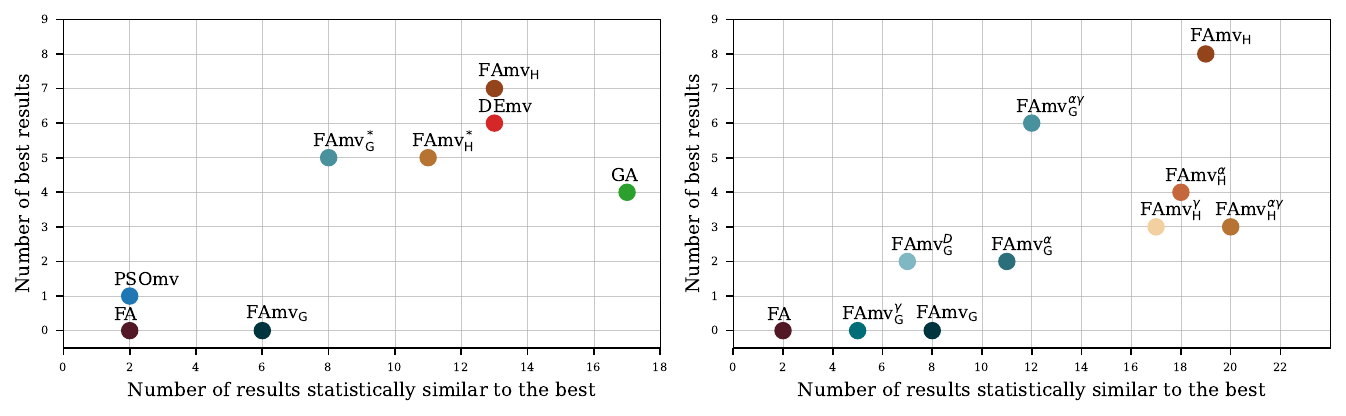}
    \caption{Overall performance of MVO-based methods on CEC benchmark functions: baseline algorithms (left) and FA-based variants (right). The x-axis shows the number of results statistically similar to the best, and the y-axis shows the number of best results. Algorithms in the top-right achieve the best performance. See Tables\ref{tab:Results_CEC2013} and \ref{tab:Ablation_Results_CEC2013} for details.}
    \label{fig:performance_overall}
\end{figure}

\begin{table*}[h!]
\centering
\small
\renewcommand{\arraystretch}{0.85}
\setlength{\tabcolsep}{5pt}
\caption{Performance of the mixed-variable optimization methods evaluated on the 28 CEC benchmark functions. Performance is evaluated in terms of the absolute error (AE). The best value (lowest average) is indicated with (*), whereas statistically similar results are highlighted in \textbf{bold}.}
\label{tab:Results_CEC2013}

\begin{tabular}{l l c c c c c c c c}
\toprule
\textbf{Function} & \textbf{Statistic} & \textbf{PSOmv} & \textbf{DEmv} & \textbf{GA} & \textbf{FA} & \textbf{FAmv$_\mathrm{H}$} & \textbf{FAmv$^{*}_\mathrm{H}$} & \textbf{FAmv$_\mathrm{G}$} & \textbf{FAmv$^{*}_\mathrm{G}$} \\
\midrule

\multicolumn{10}{l}{\textbf{Unimodal functions (F1--F5)}} \\
\midrule

\rowcolor{myg} 
F1  & Mean & 1.19e+04 & \textbf{5.76e+02} & 1.93e+03 & 1.71e+05 & \textbf{4.81e+02} & \textbf{4.47e+02}* & \textbf{5.16e+02} & \textbf{4.79e+02} \\
\rowcolor{myg} 
& STD  & 3.03e+03 & \textbf{2.13e+02} & 4.95e+02 & 1.92e+04 & \textbf{5.60e+01} & \textbf{8.43e+01}* & \textbf{5.69e+01} & \textbf{7.92e+01} \\
                     
{F2}  & Mean & 3.87e+08 & 3.31e+08 & 1.25e+08 & 7.10e+09 & \textbf{2.43e+07}* & \textbf{2.92e+07} & \textbf{3.08e+07} & \textbf{3.46e+07} \\
& STD  & 1.26e+08 & 4.92e+07 & 3.04e+07 & 1.98e+09 & \textbf{5.69e+06}* & \textbf{8.08e+06} & \textbf{8.63e+06} & \textbf{8.99e+06} \\

\rowcolor{myg}
{F3}  & Mean & 1.49e+11 & 4.77e+09 & 2.39e+10 & 1.60e+21 & \textbf{1.63e+09}* & \textbf{2.07e+09} & \textbf{2.25e+09} & 4.14e+09 \\
\rowcolor{myg}
& STD  & 2.67e+10 & 2.06e+09 & 5.05e+09 & 3.08e+21 & \textbf{5.47e+08}* & \textbf{8.97e+08} & \textbf{1.23e+09} & 4.55e+09 \\

{F4}  & Mean & 1.77e+05 & 1.43e+05 & \textbf{7.18e+04} & 5.15e+06 & \textbf{5.75e+04}* & \textbf{6.79e+04} & 9.68e+04 & 1.12e+05 \\
& STD  & 2.41e+04 & 1.44e+04 & \textbf{1.05e+04} & 1.30e+07 & \textbf{5.46e+03}* & \textbf{1.26e+04} & 1.95e+04 & 2.99e+04 \\

\rowcolor{myg}
{F5}  & Mean & 3.91e+03 & 8.50e+02 & 2.56e+03 & 2.78e+05 & \textbf{4.50e+02}* & 6.87e+02 & \textbf{5.52e+02} & 8.64e+02 \\
\rowcolor{myg}
& STD  & 1.76e+03 & 6.01e+02 & 4.32e+03 & 8.13e+04 & \textbf{5.97e+01}* & 1.15e+02 & \textbf{8.21e+01} & 1.41e+02 \\

\midrule
\multicolumn{10}{l}{\textbf{Multimodal functions (F6--F20)}} \\
\midrule

\rowcolor{myg}
{F6}  & Mean & 7.89e+02 & \textbf{6.87e+01}* & 2.00e+02 & 3.34e+04 & 1.65e+02 & 2.04e+02 & 1.57e+02 & 1.71e+02 \\
\rowcolor{myg}
& STD  & 2.23e+02 & \textbf{7.72e+00}* & 4.80e+01 & 6.74e+03 & 3.11e+01 & 3.96e+01 & 5.45e+01 & 5.43e+01 \\

{F7}  & Mean & 5.62e+05 & \textbf{3.31e+05} & \textbf{3.02e+05}* & 3.68e+10 & \textbf{3.02e+05} & \textbf{3.27e+05} & 4.27e+05 & 4.60e+05 \\
& STD  & 4.89e+04 & \textbf{2.61e+04} & \textbf{3.13e+04}* & 4.54e+10 & \textbf{3.18e+04} & \textbf{4.89e+04} & 5.82e+04 & 7.19e+04 \\

\rowcolor{myg}
{F8}  & Mean & \textbf{2.12e+01} & \textbf{2.12e+01} & \textbf{2.12e+01} & \textbf{2.11e+01} & \textbf{2.12e+01} & \textbf{2.12e+01} & \textbf{2.12e+01} & \textbf{2.12e+01}* \\
\rowcolor{myg}
& STD  & \textbf{3.00e-02} & \textbf{5.00e-02} & \textbf{4.00e-02} & \textbf{5.00e-02} & \textbf{4.00e-02} & \textbf{3.00e-02} & \textbf{4.00e-02} & \textbf{7.00e-02}* \\

{F9}  & Mean & 7.41e+01 & 6.99e+01 & \textbf{4.64e+01} & 8.56e+01 & \textbf{4.82e+01} & \textbf{4.63e+01}* & 5.16e+01 & 5.06e+01 \\
& STD  & 1.23e+00 & 1.49e+00 & \textbf{5.91e+00} & 1.67e+00 & \textbf{3.84e+00} & \textbf{3.54e+00}* & 3.50e+00 & 3.41e+00 \\

\rowcolor{myg}
{F10} & Mean & 3.00e+03 & \textbf{9.93e+01}* & 7.53e+02 & 2.75e+04 & \textbf{1.03e+02} & \textbf{1.31e+02} & \textbf{1.18e+02} & 1.83e+02 \\
\rowcolor{myg}
& STD  & 6.07e+02 & \textbf{3.51e+01}* & 2.39e+02 & 4.31e+03 & \textbf{1.48e+01} & \textbf{3.08e+01} & \textbf{2.12e+01} & 4.34e+01 \\

{F11} & Mean & 4.13e+02 & \textbf{9.17e+01}* & \textbf{1.49e+02} & 2.74e+03 & 6.84e+02 & 6.17e+02 & 7.77e+02 & 7.65e+02 \\
& STD  & 3.71e+01 & \textbf{1.33e+01}* & \textbf{1.98e+01} & 3.83e+02 & 4.72e+01 & 5.30e+01 & 8.75e+01 & 1.14e+02 \\

\rowcolor{myg}
{F12} & Mean & 6.45e+02 & \textbf{4.14e+02} & \textbf{3.79e+02}* & 2.46e+03 & 6.04e+02 & 5.59e+02 & 6.32e+02 & 5.94e+02 \\
\rowcolor{myg}
& STD  & 5.58e+01 & \textbf{1.49e+01} & \textbf{3.54e+01}* & 3.38e+02 & 4.78e+01 & 6.24e+01 & 6.59e+01 & 7.16e+01 \\

{F13} & Mean & 6.09e+02 & \textbf{4.02e+02}* & \textbf{4.11e+02} & 2.25e+03 & 5.80e+02 & 5.79e+02 & 5.99e+02 & 6.19e+02 \\
& STD  & 4.81e+01 & \textbf{1.66e+01}* & \textbf{3.04e+01} & 3.08e+02 & 3.43e+01 & 4.27e+01 & 4.22e+01 & 7.71e+01 \\

\rowcolor{myg}
{F14} & Mean & \textbf{2.00e-02}* & \textbf{2.00e-02} & \textbf{4.00e-02} & 8.95e+03 & 5.41e+03 & 3.93e+03 & 4.23e+03 & 3.30e+03 \\
\rowcolor{myg}
& STD  & \textbf{1.00e-02}* & \textbf{2.00e-02} & \textbf{4.00e-02} & 7.19e+02 & 8.71e+02 & 1.08e+03 & 1.17e+03 & 1.96e+03 \\

{F15} & Mean & 9.11e+03 & 9.47e+03 & 7.03e+03 & 1.05e+04 & 8.46e+03 & 6.94e+03 & 7.09e+03 & \textbf{5.52e+03}* \\
& STD  & 7.04e+02 & 5.89e+02 & 1.34e+03 & 7.18e+02 & 8.05e+02 & 1.16e+03 & 1.01e+03 & \textbf{1.10e+03}* \\

\rowcolor{myg}
{F16} & Mean & 3.80e+00 & 3.76e+00 & 3.79e+00 & \textbf{3.42e+00} & 3.79e+00 & 3.74e+00 & 3.75e+00 & \textbf{3.42e+00}* \\
\rowcolor{myg}
& STD  & 3.80e-01 & 2.80e-01 & 2.90e-01 & \textbf{4.00e-01} & 2.90e-01 & 3.00e-01 & 3.80e-01 & \textbf{3.60e-01}* \\

{F17} & Mean & 7.96e+02 & \textbf{1.99e+02}* & \textbf{3.55e+02} & 5.39e+03 & 6.42e+02 & 6.58e+02 & 6.69e+02 & 7.10e+02 \\
& STD  & 1.10e+02 & \textbf{2.63e+01}* & \textbf{4.11e+01} & 3.88e+02 & 3.88e+01 & 5.87e+01 & 3.33e+01 & 5.91e+01 \\

\rowcolor{myg}
{F18} & Mean & 9.03e+02 & \textbf{1.54e+02}* & \textbf{3.30e+02} & 7.29e+03 & 7.29e+02 & 8.47e+02 & 7.75e+02 & 7.82e+02 \\
\rowcolor{myg}
& STD  & 1.38e+02 & \textbf{2.08e+01}* & \textbf{4.84e+01} & 7.63e+02 & 6.18e+01 & 9.59e+01 & 5.71e+01 & 1.18e+02 \\

{F19} & Mean & 3.45e+03 & 4.65e+01 & 7.37e+01 & 2.08e+07 & \textbf{3.48e+01}* & 4.96e+01 & 4.91e+01 & 5.05e+01 \\
& STD  & 2.49e+03 & 2.21e+00 & 2.34e+01 & 1.22e+07 & \textbf{1.49e+00}* & 3.59e+00 & 2.73e+00 & 4.46e+00 \\

\rowcolor{myg}
{F20} & Mean & 2.45e+01 & 2.39e+01 & \textbf{2.31e+01} & 2.50e+01 & 2.36e+01 & \textbf{2.30e+01}* & 2.41e+01 & 2.41e+01 \\
\rowcolor{myg}
& STD  & 4.80e-01 & 2.60e-01 & \textbf{7.40e-01} & 0.00e+00 & 8.20e-01 & \textbf{5.40e-01}* & 6.20e-01 & 5.90e-01 \\

\midrule
\multicolumn{10}{l}{\textbf{Composition functions (F21--F28)}} \\
\midrule

\rowcolor{myg}
{F21} & Mean & 3.05e+03 & 7.03e+02 & 9.78e+02 & 1.22e+04 & \textbf{5.60e+02} & \textbf{5.49e+02}* & 6.07e+02 & \textbf{6.79e+02} \\
\rowcolor{myg}
& STD  & 4.54e+02 & 5.82e+01 & 3.19e+02 & 1.30e+03 & \textbf{8.36e+00} & \textbf{9.66e+00}* & 2.14e+02 & \textbf{3.62e+02} \\

{F22} & Mean & 7.00e-02 & \textbf{3.00e-02} & \textbf{3.00e-02}* & 1.26e+04 & 1.08e+04 & 1.04e+04 & 8.31e+03 & 7.08e+03 \\
& STD  & 1.00e-01 & \textbf{4.00e-02} & \textbf{4.00e-02}* & 6.44e+02 & 1.10e+03 & 1.20e+03 & 1.62e+03 & 2.11e+03 \\

\rowcolor{myg}
{F23} & Mean & 9.66e+03 & 9.98e+03 & \textbf{7.14e+03} & 1.21e+04 & 1.06e+04 & 9.50e+03 & 8.71e+03 & \textbf{6.82e+03}* \\
\rowcolor{myg}
& STD  & 8.00e+02 & 7.72e+02 & \textbf{2.01e+03} & 6.86e+02 & 1.02e+03 & 9.89e+02 & 1.23e+03 & \textbf{1.18e+03}* \\

{F24} & Mean & 3.77e+02 & 3.57e+02 & 3.76e+02 & 1.71e+03 & \textbf{3.19e+02}* & 3.72e+02 & 3.69e+02 & 3.63e+02 \\
& STD  & 6.75e+00 & 4.34e+00 & 1.26e+02 & 4.22e+02 & \textbf{9.44e+00}* & 1.02e+02 & 1.87e+01 & 1.61e+01 \\

\rowcolor{myg}
{F25} & Mean & 3.91e+02 & \textbf{3.77e+02} & \textbf{3.46e+02}* & 7.70e+02 & 4.06e+02 & 4.11e+02 & 4.31e+02 & 4.15e+02 \\
\rowcolor{myg}
& STD  & 1.11e+01 & \textbf{4.40e+00} & \textbf{1.16e+01}* & 6.41e+01 & 1.69e+01 & 1.43e+01 & 1.57e+01 & 1.87e+01 \\

{F26} & Mean & 4.71e+02 & 4.51e+02 & \textbf{4.12e+02} & 2.20e+03 & 4.90e+02 & 4.30e+02 & 5.28e+02 & \textbf{3.29e+02}* \\
& STD  & 1.63e+01 & 1.41e+02 & \textbf{1.26e+01} & 1.53e+03 & 3.82e+02 & 1.06e+02 & 3.79e+02 & \textbf{6.26e+01}* \\

\rowcolor{myg}
{F27} & Mean & 2.05e+03 & 1.94e+03 & \textbf{1.49e+03} & 6.28e+03 & \textbf{1.47e+03}* & 1.61e+03 & 1.88e+03 & 1.79e+03 \\
\rowcolor{myg}
& STD  & 7.64e+01 & 3.88e+01 & \textbf{1.18e+02} & 1.75e+03 & \textbf{3.53e+02}* & 2.47e+02 & 1.41e+02 & 1.37e+02 \\

{F28} & Mean & 4.89e+03 & 4.74e+03 & \textbf{2.61e+03} & 4.64e+04 & 5.51e+03 & \textbf{1.45e+03}* & 5.77e+03 & 4.61e+03 \\
& STD  & 6.01e+02 & 6.52e+02 & \textbf{4.56e+02} & 9.05e+04 & 5.21e+02 & \textbf{1.19e+02}* & 7.57e+02 & 1.12e+03 \\

\midrule
\rowcolor{mylblue}
{\textbf{Count results }} 
&
& \textbf{2} \textbf{(1)}
& \textbf{13} \textbf{(6)}
& \textbf{17} \textbf{(4)}
& \textbf{2} \textbf{(0)}
& \textbf{13} \textbf{(7)}
& \textbf{11} \textbf{(5)}
& \textbf{6} \textbf{(0)}
& \textbf{8} \textbf{(5)} \\
\bottomrule
\end{tabular}
\end{table*}

\begin{figure}
    \centering
    \includegraphics[width=0.8\linewidth]{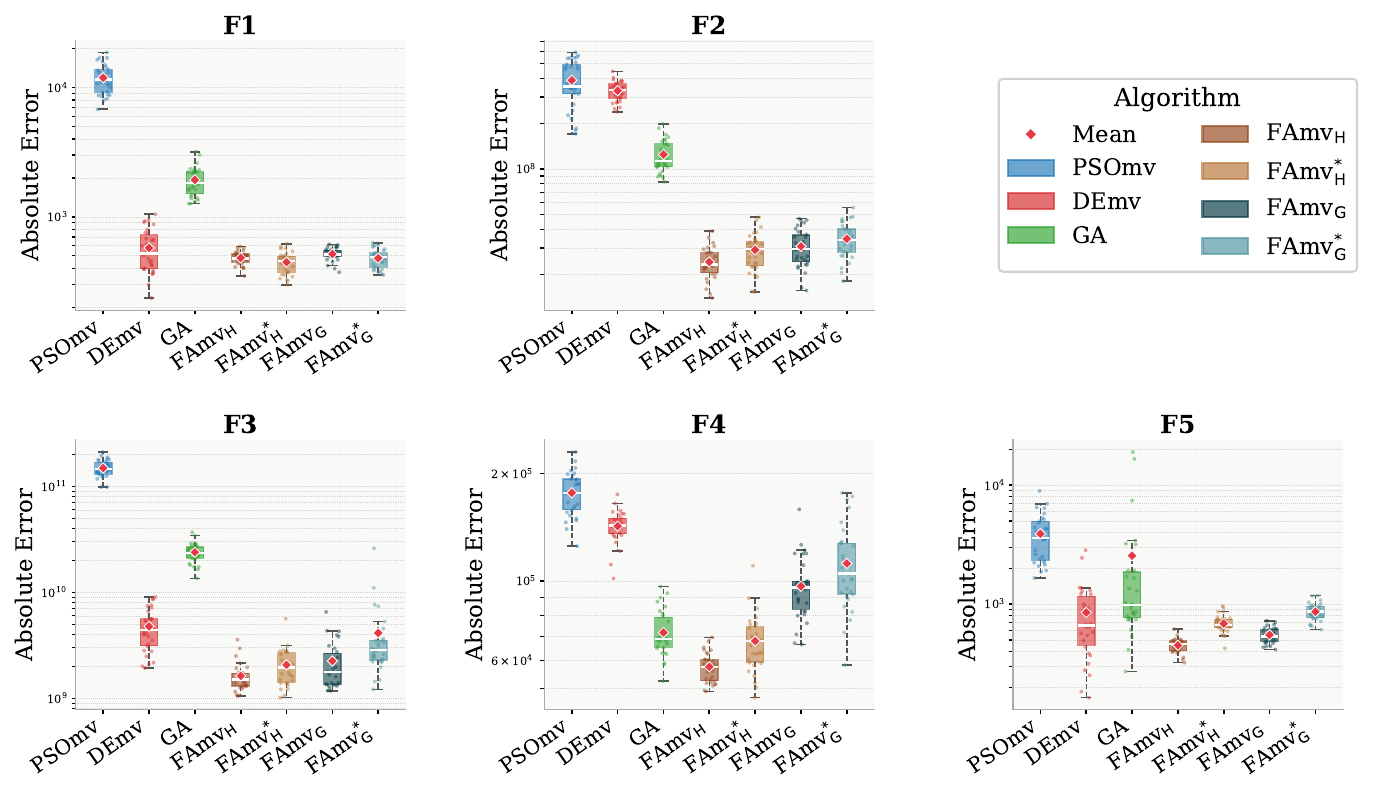}
    \caption{Performance in terms of the absolute error (the lower the better) scored by the seven mixed-variable optimization methods on the CEC benchmark for the \textbf{unimodal functions}.}
    \label{fig:bp1}
\end{figure}

\begin{figure}
    \centering
    \includegraphics[width=1\linewidth]{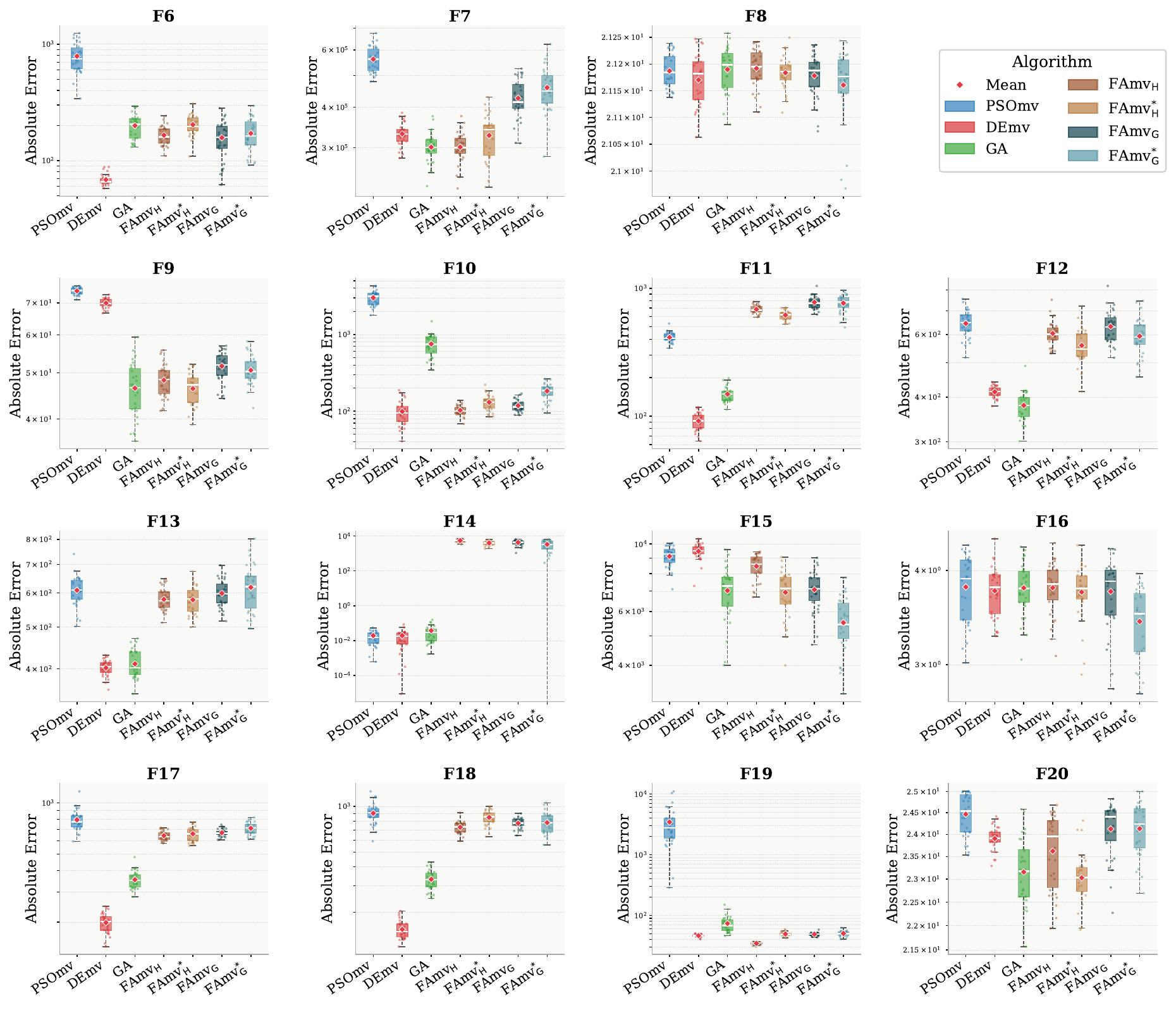}
    \caption{Performance in terms of the absolute error (the lower the better) scored by the seven mixed-variable optimization methods on the CEC benchmark for the \textbf{multimodal functions}.}
    \label{fig:bp2}
\end{figure}

\begin{figure} 
    \centering
    \includegraphics[width=1\linewidth]{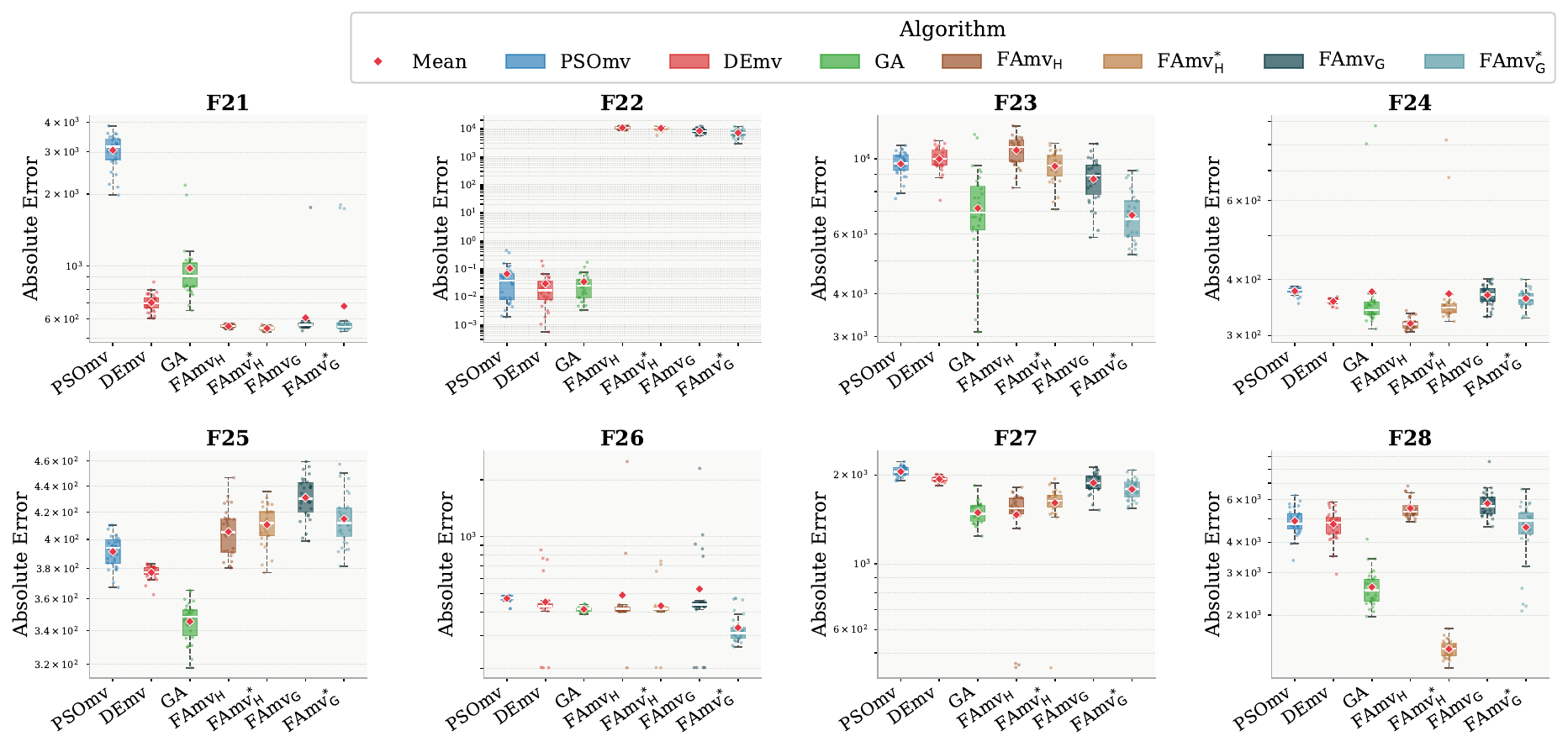}
    \caption{
    Performance in terms of the absolute error (the lower the better) scored by the seven MVO methods on \textbf{composite functions} in the CEC benchmark.}
    \label{fig:bp3}
\end{figure}

\subsection{Results on Benchmark Functions} \label{subsec:results_benchmark_CEC2013}

This section analyzes the results on the CEC2013 benchmark obtained by our FAmv algorithm and compares its performance with a set of mixed-variable reference approaches (see Section~\ref{subsec:setup_baselines}). The results of this comparison are summarized in Table~\ref{tab:Results_CEC2013} and Figure~\ref{fig:performance_overall} (left). Figures of performance with specific results for separate problem categories are also presented: unimodal (Figures~\ref{fig:bp1}), multimodal (Figures~\ref{fig:bp2}), and composition (Figures~\ref{fig:bp3}) functions, providing a more detailed analysis of algorithm behavior across different landscape characteristics, and the respective convergence plots are presented in the additional material (Appendix~\ref{app:convergence_plots}). 

As shown in Figure~\ref{fig:performance_overall} (left) and Table~\ref{tab:Results_CEC2013}, FAmv$_\mathrm{H}$, DEmv, and GA exhibit the strongest overall performance in terms of optimality and robustness. It can be observed that the proposed FAmv variants, and particularly the FAmv$_\mathrm{H}$ variant based on the Hamming distance, achieve lower absolute error (AE) values than the competing approaches in terms of optimality (see at the bottom of Table~\ref{tab:Results_CEC2013}). Specifically, FAmv$_\mathrm{H}$ obtained the best results for 7 functions (F2, F3, F4, F5, F19, F24 and F27), whereas DEmv performed better in six functions (F6, F10, F11, F13, F17 and F18) and GA in four cases (F7, F12, F22 and F25). In terms of robustness, measured as the number of functions for which an algorithm is statistically equivalent to the best, GA performed best on 17 functions, followed by DEmv and FAmv with 13 each. These overall performances in terms of optimality and robustness are shown in Figure~\ref{fig:performance_overall}(left).

\textbf{Unimodal functions (F1–F5)}. In particular, FAmv$_\mathrm{H}$ achieves the best results on functions F2, F3, F4, and F5. These observations are also supported by the convergence curves shown in Figure~\ref{fig:conv1}. For functions F1 and F2, the proposed FAmv variants exhibit faster convergence and reach lower final errors than the other algorithms. Furthermore, the boxplots presented in Figure~\ref{fig:bp1} illustrate the stability of the proposed approaches. Compared to other algorithms, the FAmv variants exhibit lower solution dispersion, indicating greater stability across independent runs. For functions F1 and F5, the proposed methods exhibit reduced variability, whereas for F2, F3, and F4, the Hamming-based FAmv variants maintain consistently stable performance.

\textbf{Multimodal functions (F6–F20)}. The results reported in Table~\ref{tab:Results_CEC2013} show a more heterogeneous behavior across the compared algorithms. Nevertheless, the proposed FAmv variants remain competitive in several functions of this category. In particular, the proposed approaches achieve the best results on functions F8, F9, F15, F16, F19, and F20. For these functions, the FAmv variants outperform all baseline algorithms in terms of mean error. Furthermore, for functions such as F7 and F10, the results remain statistically similar to those of the best-performing algorithms, indicating competitive performance compared to the considered state-of-the-art approaches. The boxplots shown in Figure~\ref{fig:bp2} provide additional insight into the distribution of the obtained solutions. For several functions, including F8, F10, and F19, the FAmv variants exhibited low solution dispersion across independent runs. This indicates consistent performance across multiple runs on these multimodal problems.

\textbf{Composition functions (F21–F28)}. We observed  
strong performance of the proposed FAmv-based variants. In particular, the proposed methods achieve the best results on six out of the eight composition functions, namely F23, F24, F26, F27, and F28. In addition to achieving the best mean performance on most of these functions, the obtained results also exhibit low dispersion across independent runs, as reflected by the standard deviation values and illustrated in the boxplots in Figure~\ref{fig:bp3}.

\subsection{Results on Engineering Design Problems} \label{subsec:results_engineering_problems}

\begin{figure} [htb!]
    \centering
    \includegraphics[width=0.8\linewidth]{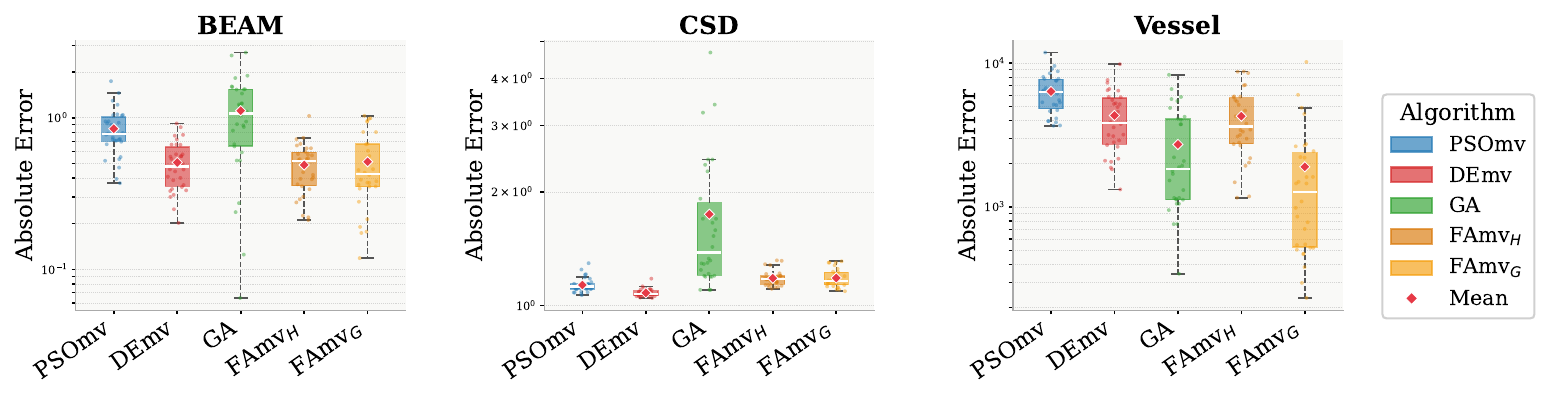}
    \includegraphics[width=0.8\linewidth]{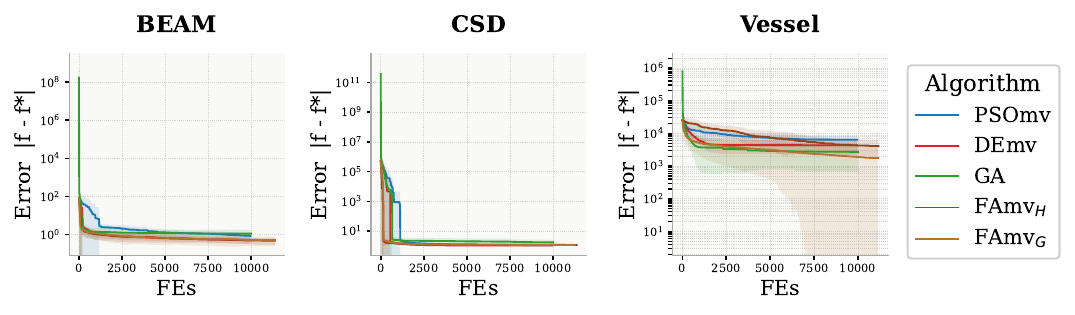}
    \caption{Performance of MVO methods on the three engineering design problems (BEAM, CSD, and Vessel), measured by absolute error (top) and convergence curves (bottom). Lower values indicate better performance.
    }
    \label{fig:eng_results}
\end{figure}

To further investigate the performance of the proposed algorithm on real-world mixed-variable optimization problems, three engineering design problems of varying complexity are considered. These results are summarized in Table~\ref{tab:Results_eng} and Figure~\ref{fig:eng_results}.

From Table~\ref{tab:Results_eng}, we observe that the proposed FA variants achieve the best performance on two out of the three considered problems, namely the BEAM and Vessel design problems. For the BEAM design problem, the FAmv variant, FAmv$_\mathrm{H}$, achieves the best result. In addition, the boxplot in Figure~\ref{fig:eng_results} illustrates a very small dispersion of the obtained solutions, indicating a stable optimization behavior across independent runs. For the Pressure Vessel problem, the mixed-variable Firefly FAmv$_\mathrm{G}$ achieves the best performance among the compared algorithms. For the CSD problem, DEmv achieves the best result, while the FAmv variants obtain slightly higher mean errors but remain close in performance with low variability across runs, as indicated by their standard deviation values. These results further demonstrate the competitiveness of the proposed mixed-variable Firefly variants in real-world engineering design problems.

\begin{table*}[ht!]
\centering
\renewcommand{\arraystretch}{1.3}
\setlength{\tabcolsep}{8pt}
\caption{Performance of MVO methods on the three engineering design problems, measured by the absolute error (AE). The best value (lowest) is indicated with (*), and statistically similar results are highlighted in \textbf{bold}.
}
\label{tab:Results_eng}

\begin{tabular}{lccccc}
\toprule
\textbf{Function} & \textbf{PSOmv} & \textbf{DEmv} & \textbf{GA} & \textbf{FAmv$_\mathrm{H}$} & \textbf{FAmv$_\mathrm{G}$} \\
\midrule

\rowcolor{myg}
\textbf{BEAM}
& 0.85 {\scriptsize $\pm$ 0.30}
& \textbf{0.51} {\scriptsize $\pm$ \textbf{0.18}}
& 1.11 {\scriptsize $\pm$ 0.64}
& \textbf{0.49}* {\scriptsize $\pm$ \textbf{0.18}}
& \textbf{0.51} {\scriptsize $\pm$ \textbf{0.27}} \\

\textbf{CSD}
& 1.13 {\scriptsize $\pm$ 0.05}
& \textbf{1.08}* {\scriptsize $\pm$ \textbf{0.03}}
& 1.74 {\scriptsize $\pm$ 0.81}
& 1.18 {\scriptsize $\pm$ 0.05}
& 1.18 {\scriptsize $\pm$ 0.06} \\

\rowcolor{myg}
\textbf{Vessel}
& 6363.68 {\scriptsize $\pm$ 1978.77}
& 4331.97 {\scriptsize $\pm$ 2040.48}
& \textbf{2724.67} {\scriptsize $\pm$ \textbf{2055.74}}
& 4283.55 {\scriptsize $\pm$ 2164.54}
& \textbf{1903.08}* {\scriptsize $\pm$ \textbf{2090.71}} \\

\bottomrule
\end{tabular}
\end{table*}

\subsection{Ablation Study of the Proposed FAmv Components} \label{subsec:ablation_study}

This section presents an ablation study to analyze the contribution of the main components of FAmv, namely the mixed-variable movement mechanism and the parameter adaptation strategy as detailed in Sections~\ref{subsec:famv_movement} and~\ref{subsec:famv_adaptation}, respectively. In total, eight FA-based variants were tested to investigate the impact of certain design choices. Four FA-variants based on mixed Euclidean-Hamming (FAmv$_\mathrm{H}$, FAmv$^{\alpha}_\mathrm{H}$, FAmv$^{\alpha\gamma}_\mathrm{H}$ and FAmv$^{\gamma}_\mathrm{H}$) and four variants based on the Gower distance (FAmv$_\mathrm{G}$, FAmv$^{\alpha}_\mathrm{G}$, FAmv$^{\alpha\gamma}_\mathrm{G}$ and FAmv$^{\gamma}_\mathrm{G}$). In these definitions, the superscript indicates the parameter or parameters that are adapted. For example, ${\alpha\gamma}$ indicates both parameters are adapted simultaneously, whereas the superscript ${\alpha}$ indicates that only ${\alpha}$ is adapted and therefore ${\gamma}$ remains fixed. Additionally, the original FA algorithm is considered as a baseline to assess the necessity of the proposed mixed-variable adaptations.

The results of this study are summarized in Figure~\ref{fig:performance_overall} (right) and Table~\ref{tab:Ablation_Results_CEC2013} (in the additional material). Additionally, detailed performance and convergence analyses for separate problem categories, unimodal (Figures~\ref{fig:ab_bp1} and~\ref{fig:ab_conv1}), multimodal (Figures~\ref{fig:ab_bp2} and~\ref{fig:ab_conv2}), and composition (Figures~\ref{fig:ab_bp3} and~\ref{fig:ab_conv3}), are provided in the additional material (Appendix~\ref{app:results_ablation}).

First, it can be observed that all the proposed FA-variants consistently outperform the classical FA algorithm on the CEC2013 mixed-variable benchmark. The proposed FA variants achieve strictly better results on almost all functions, except for function F8, where all algorithms are statistically similar. The results indicate that the Hamming-based variants provide consistently strong performance across most function categories, regardless of whether fixed or adaptive parameters are used. Interestingly, the variant using fixed control parameters remains frequently among the best or statistically competitive approaches, suggesting that the proposed mixed-variable movement mechanism alone contributes significantly to performance improvements. At the same time, variants employing adaptive parameters FAmv$^{\alpha, \gamma}_\mathrm{H}$ also demonstrate strong performance on more complex multimodal and composition functions, indicating that parameter adaptation further enhances performance in more challenging landscapes. Regarding the Gower-based distance, the most competitive variants are those that employ adaptive control parameters, which appear more frequently among the best-performing algorithms than those using fixed parameters, highlighting the importance of parameter adaptation when using more general mixed-variable distance measures.

These results can be explained by the role of distance in the attractiveness term of the original FA, which modulates interactions via an exponential decay function. Since the Gower distance is bounded in $[0,1]$, the exponential term remains relatively large, leading to consistently strong attraction between individuals and increased exploitation of the search space. In contrast, the Hamming-based distance can take larger values, resulting in stronger decay of the attractiveness term and enabling a wider range of movement behaviors.
This difference in scale provides a more effective modulation of the attraction mechanism when using the Hamming-based distance, leading to more stable and competitive performance across benchmark functions. Conversely, variants based on the Gower distance may require careful tuning of the $\gamma$ parameter to compensate for its bounded nature and achieve a similar balance between exploration and exploitation.



\section{Discussion} \label{sec:discussion}

The experimental results reveal distinct behaviors of the proposed algorithm among benchmark function categories. In particular, the proposed FA variants show strong dominance over unimodal and composition functions, whereas performance is more heterogeneous on classical multimodal functions. 

For unimodal functions, the optimization landscape typically contains a single global optimum and relatively smooth gradients. In such cases, strong exploitation capabilities are generally sufficient to guide the search process toward the optimum. The proposed Firefly Algorithm (FA) variants, especially those based on the Hamming distance, exhibit strong exploratory behavior when individuals are distant in the search space, while maintaining efficient local exploitation as they become closer. This balance allows the algorithm to efficiently converge toward the global optimum, which explains the strong performance observed on these functions.

In contrast, classical multimodal functions contain many local optima distributed across the search space. In such landscapes, maintaining a careful balance between exploration and exploitation becomes significantly more challenging. The proposed variants may sometimes operate in two extreme regimes depending on the distance formulation. Variants using the Hamming distance tend to produce stronger exploration because of potentially large distance values, while variants relying on the Gower distance often lead to stronger exploitation because the distance is normalized to $[0,1]$. This behavior can explain the more heterogeneous results observed on multimodal functions.

For composition functions, which combine multiple landscape characteristics and are typically more complex, the algorithm's exploitation capability becomes particularly important. In these cases, variants based on the Gower distance tend to perform better, as the bounded nature of the distance leads to stronger attraction between individuals and thereby promotes the exploitation of promising regions in the search space. This behavior is consistent with the experimental results for the benchmark's composition functions.

Another important aspect of the proposed approach is its sensitivity to hyperparameters. This sensitivity mainly originates from the exponential attractiveness function used in the FA. The attractiveness between individuals is modeled using an exponential decay function that depends directly on the distance between solutions and the parameter gamma, which controls the rate of decay. Due to the nature of the exponential function, small variations in the distance or in the parameter gamma can produce significant variations in the attractiveness value. As a result, the balance between exploration and exploitation may vary substantially depending on the parameter values chosen. This characteristic explains why different variants of the algorithm may exhibit different search behaviors depending on the distance formulation and parameter settings. Although adaptive parameters partially mitigate this issue by allowing the algorithm to automatically adjust its search behavior during optimization, achieving a perfectly balanced exploration-exploitation trade-off remains difficult because of the intrinsic sensitivity of the exponential decay mechanism.

\textbf{Main limitations}. Despite the encouraging results obtained in this study, some limitations should be acknowledged. One important limitation arises from the mixed-variable distance formulation, particularly when using the Hamming-based distance. Since this distance is not normalized, the continuous component of the solution may dominate the overall distance value when continuous variables span very large intervals. In such situations, the contribution of discrete variables to the distance computation may become relatively less significant, potentially affecting the algorithm's search dynamics. Another limitation concerns the difficulty of achieving a stable exploration-exploitation balance within the FA framework. Because the search dynamics heavily rely on the exponential attractiveness function, the algorithm can sometimes operate in regimes that favor either strong exploration or strong exploitation. Although the introduction of random movements controlled by the alpha parameter partially compensates for this behavior, identifying robust parameter settings remains challenging.



\section{Conclusions} \label{sec:conclusions}

In this paper, we proposed an adaptation of the Firefly Algorithm (FA) to optimization problems defined over mixed-variable search spaces. The proposed approach explicitly considers heterogeneous decision variables, including continuous, ordinal, discrete, and categorical variables. To achieve this, we introduced a mixed-variable movement mechanism and a distance-based modeling strategy that integrates both continuous and discrete components of candidate solutions into the FA's attractiveness mechanism. These adaptations allow the algorithm to operate more effectively in mixed-variable optimization settings by respecting the intrinsic structure of the search space. In addition, a parameter-adaptation strategy was incorporated to balance exploration and exploitation during the search. In particular, the proposed formulation enables joint integration of dimensions of different types, allowing the algorithm to respond to interaction structures through a unified distance-based interaction mechanism.

The experimental results demonstrate the effectiveness of the proposed approach. Extensive experiments on the CEC2013 mixed-variable benchmark show that the proposed algorithm achieves competitive performance, with several variants achieving superior results across multiple functions compared to several baseline algorithms. The results show both the accuracy and the robustness of the proposed method across different categories of benchmark functions. In addition, the proposed algorithm was evaluated on three engineering design problems, further validating its performance on real-world optimization tasks. The results obtained on these engineering problems are consistent with the overall trends observed on the synthetic benchmark, confirming the practical usefulness of the proposed approach.

Despite these encouraging results, several directions remain open for future research. In particular, future work will focus on improving the distance formulation to better balance the contributions of continuous and discrete variables, especially when continuous variables span large ranges. Another promising direction is to develop more advanced adaptive mechanisms that automatically regulate the exploration–exploitation trade-off during the optimization process. Regarding the applicability of FAmv to real-world optimization problems with mixed variables, potential application domains include clustering of mixed-variable data and AutoML (automatic parameter configuration of algorithms).

\section*{Acknowledgment}
This work was supported by the French government through the Programme d’Investissement d’Avenir (I-SITE ULNE / ANR-16-IDEX-0004 ULNE), managed by the Agence Nationale de la Recherche (No. I-KUL-22-005-ARCHIE-INFINITE); by a grant from Inserm and the French Ministry of Health within the MESSIDORE 2023 call operated by IReSP (AAP-2023-MSDR-341423); and by the European Union’s Horizon Europe Research and Innovation Programme under the Marie Skłodowska-Curie Actions (MSCA), Grant Agreement No. 101236749.

\bibliographystyle{plainnat}
\bibliography{references/references}

\appendix
\newpage
\section{FA Pseudocode}
\label{app:FA_algorithm}

The main steps of the Firefly algorithm\footnote{A minimization problem is considered; for maximization problems, it suffices to minimize $g = -f$.} are summarized in Algorithm~\ref{alg:firefly}.


\begin{algorithm}[htb!]
\DontPrintSemicolon
\setstretch{1.2}

Randomly initialize the population of $N$ fireflies, $X_1, X_2, \ldots, X_N$\;
Compute the fitness value of each firefly using $f$\;
$t \leftarrow 1$\;

\While{$t \leq MAX\_ITER$}{
    \For{$i \leftarrow 1$ \KwTo $N$}{
        \For{$j \leftarrow 1$ \KwTo $N$}{
            \If{$f(X_j) < f(X_i)$}{
                Move $X_i$ toward $X_j$ according to Eq.~\ref{eq:update_position}\;
                Compute the fitness value of the new $X_i$\;
                $t \leftarrow t + 1$\;
            }
        }
    }
}

\caption{\label{alg:firefly}Firefly algorithm}
\end{algorithm}

\begin{figure} [ht!]
    \centering
    \includegraphics[width=0.6\linewidth]{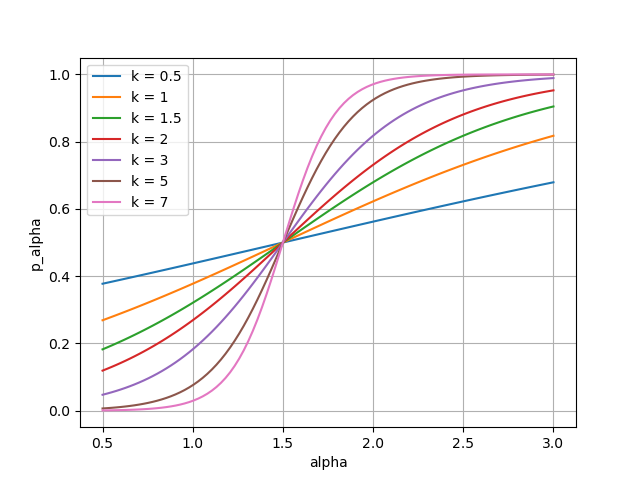}
    \caption{The influence of the control parameter $k$ on the probability function $p_\alpha$ presented in Section~\ref{subsec:famv_movement}}
    \label{fig:p_alpha}
\end{figure}

The influence of the control parameter $k$ on the probability function $p_\alpha$ in our FAmv algorithm is illustrated in Figure~\ref{fig:p_alpha}. The description of this parameter is presented in Section~\ref{subsec:famv_movement}.

\section{Convergence plots: Benchmark functions}
\label{app:convergence_plots}

\begin{figure}
    \centering
    \includegraphics[width=0.75\linewidth]{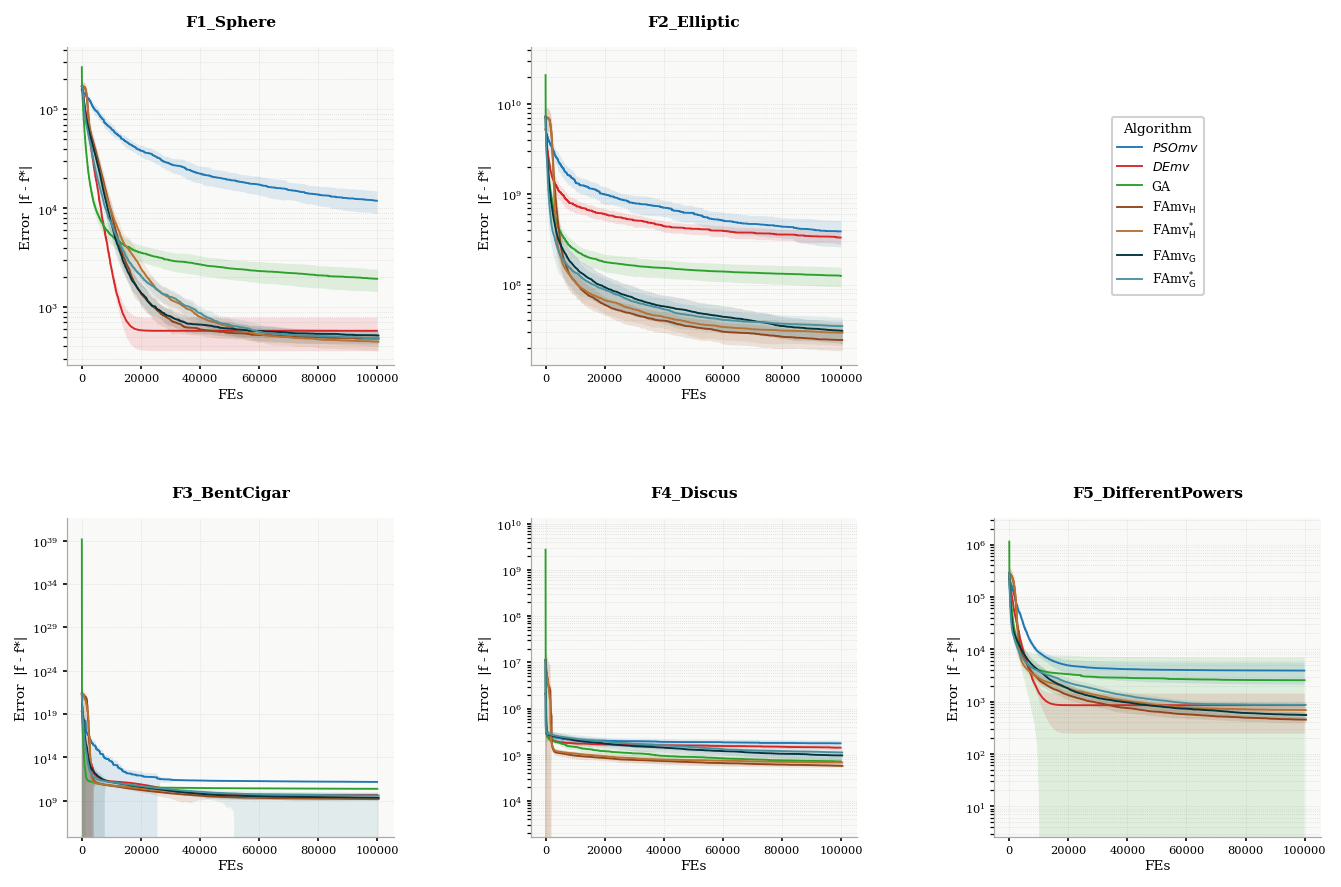}
    \caption{
    Convergence curves of the MVO methods on \textbf{unimodal functions} in the CEC benchmark.}
    \label{fig:conv1}
\end{figure}

\begin{figure}
    \centering
    \includegraphics[width=1\linewidth]{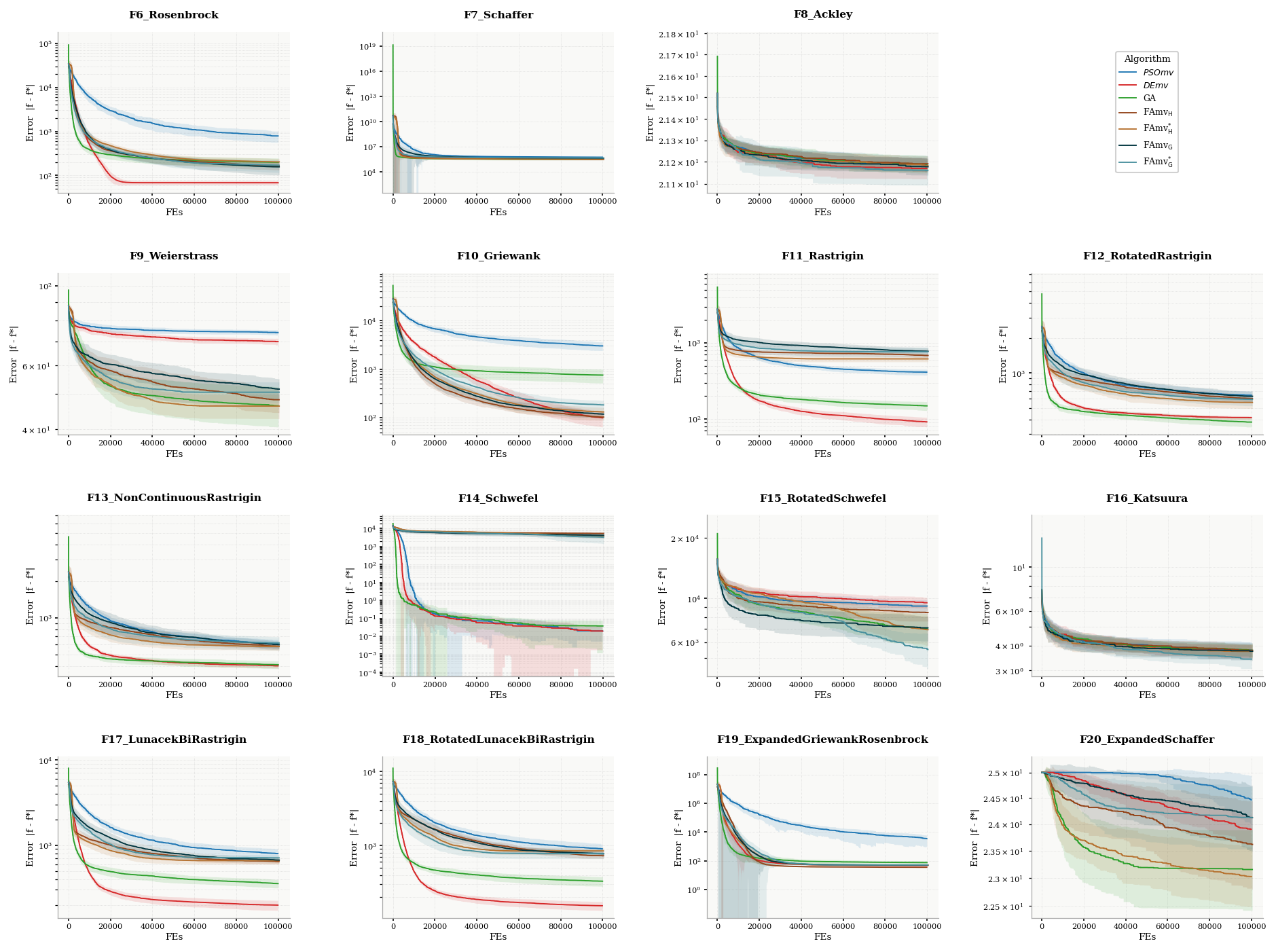}
    \caption{
    Convergence curves of the MVO methods on \textbf{multimodal functions} in the CEC benchmark.}
    \label{fig:conv2}
\end{figure}

\begin{figure}
    \centering
    \includegraphics[width=1\linewidth]{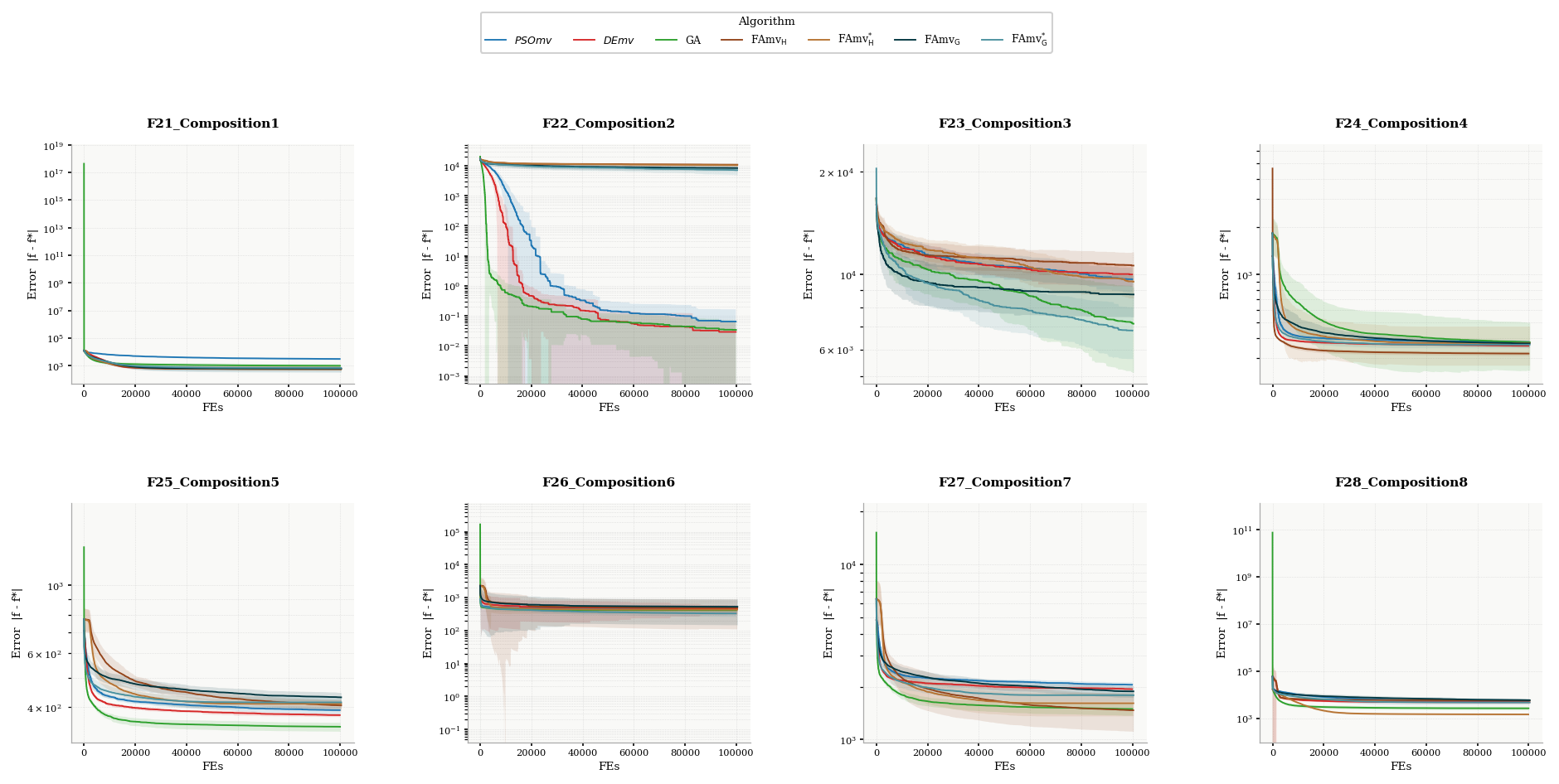}
    \caption{
    Convergence curves of the MVO methods on \textbf{composite functions} in the CEC benchmark.}
    \label{fig:conv3}
\end{figure}

This section presents complementary convergence plots of the results on the CEC2013 benchmark obtained by our FAmv algorithm, compared with a set of mixed-variable reference approaches (see Section~\ref{subsec:setup_baselines}). The results of this comparison are summarized in Table~\ref{tab:Results_CEC2013}. Figures of convergence with specific results for separate problem categories are presented: unimodal (Figure~\ref{fig:conv1}), multimodal (Figure~\ref{fig:conv2}), and composition (Figure~\ref{fig:conv3}) functions.

\section{Full results on the impact of the mixed-variable movement mechanism in FA variants}
\label{app:results_ablation}

\begin{figure} [ht!]
    \centering
    \includegraphics[width=0.75\linewidth]{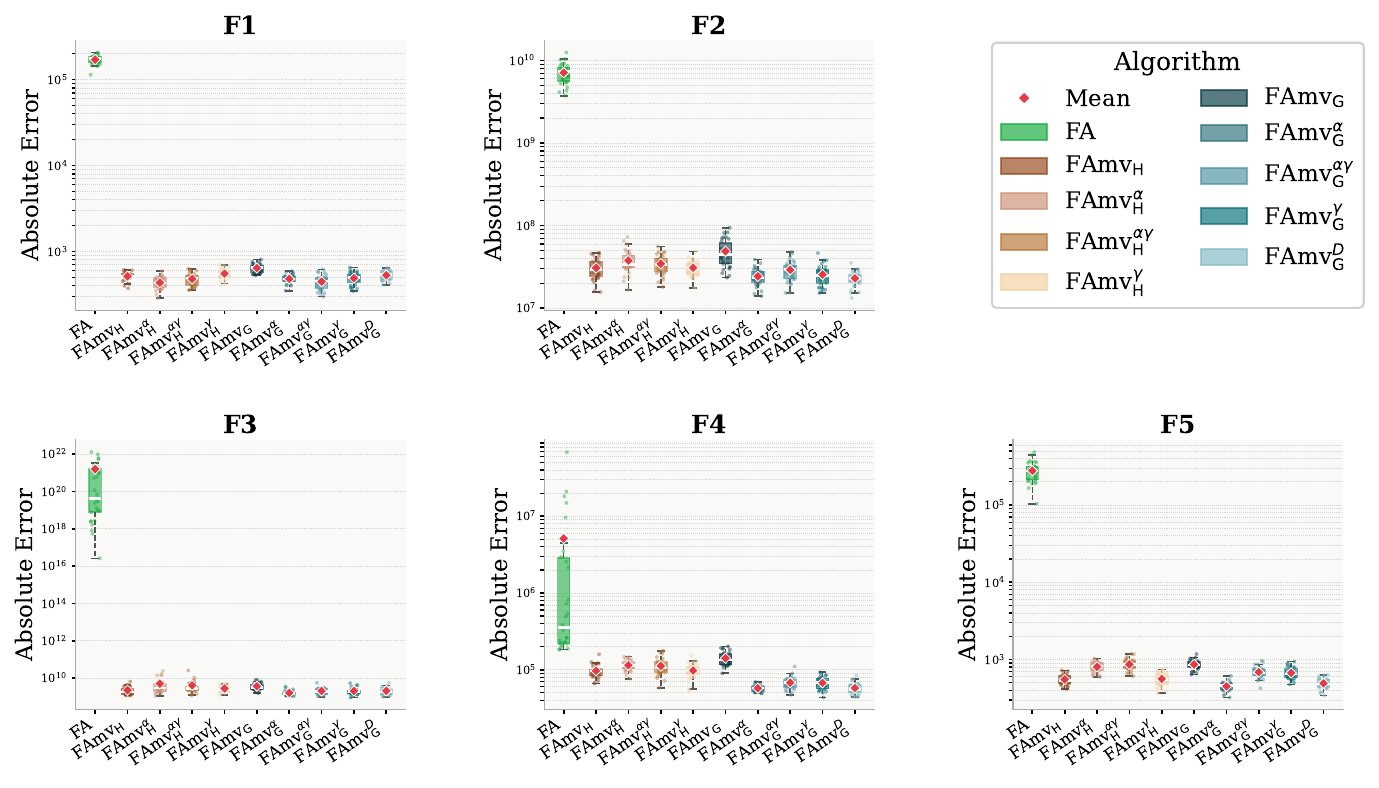}
    \caption{
    Performance scored by the FA-based variants on the CEC benchmark for the \textbf{unimodal functions}.}
    \label{fig:ab_bp1}
\end{figure}

\begin{figure} [ht!]
    \centering
    \includegraphics[width=1\linewidth]{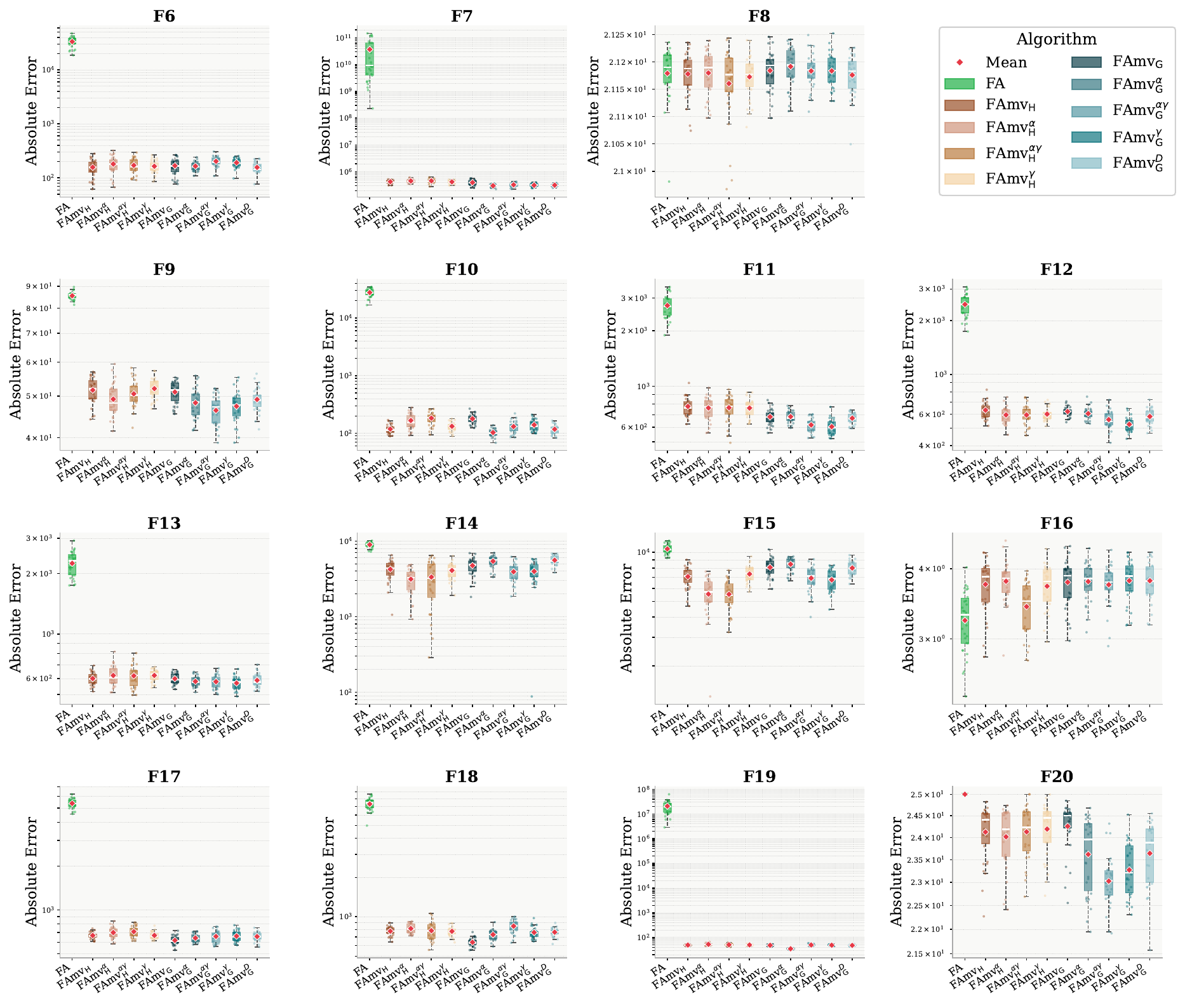}
    \caption{
    Performance scored by the FA-based variants on the CEC benchmark for the \textbf{multimodal functions}.}
    \label{fig:ab_bp2}
\end{figure}

\begin{figure} [ht!]
    \centering
    \includegraphics[width=1\linewidth]{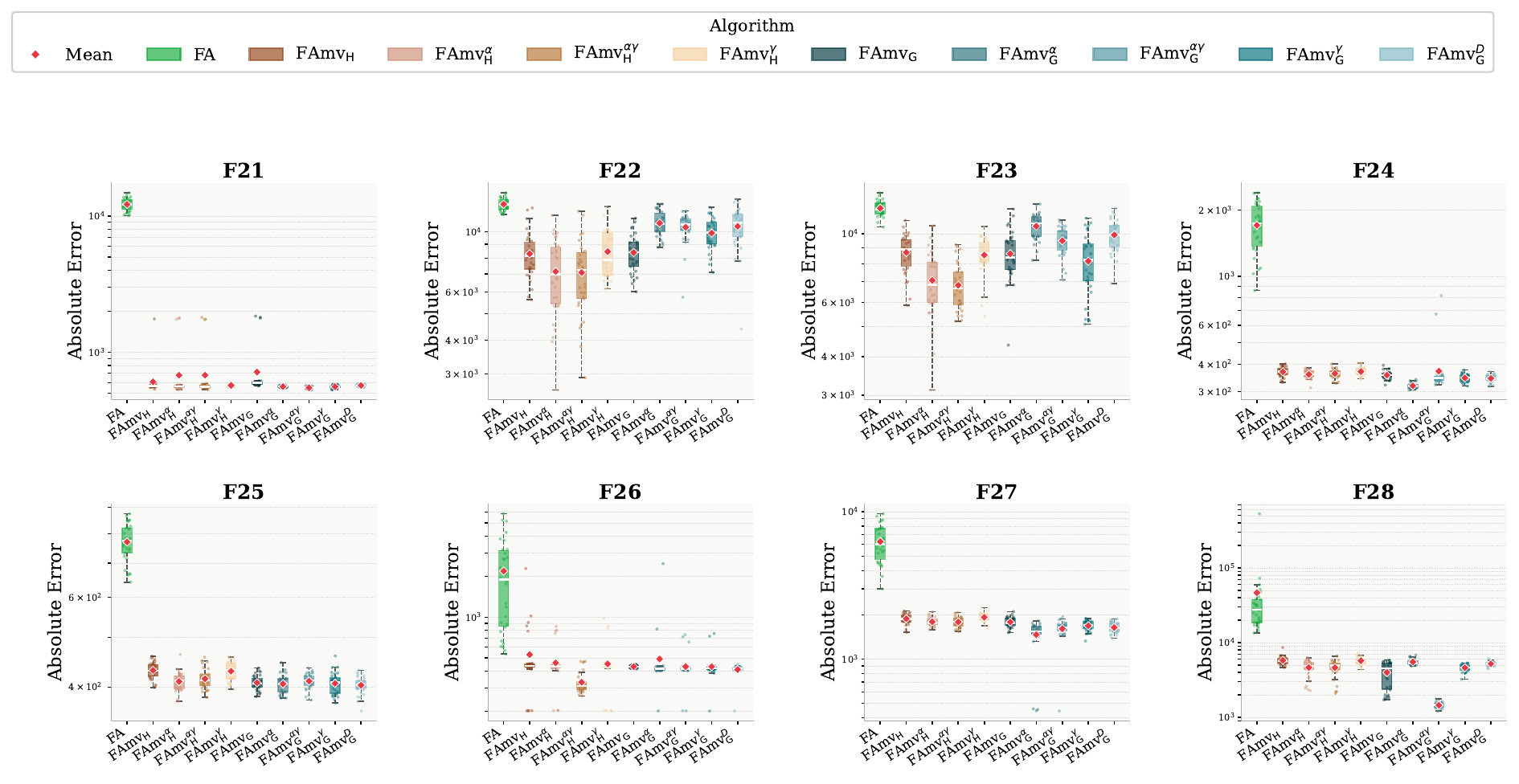}
    \caption{
    Performance scored by the FA-based variants on the CEC benchmark for the \textbf{composition functions}.}
    \label{fig:ab_bp3}
\end{figure}

\begin{figure}
    \centering
    \includegraphics[width=0.75\linewidth]{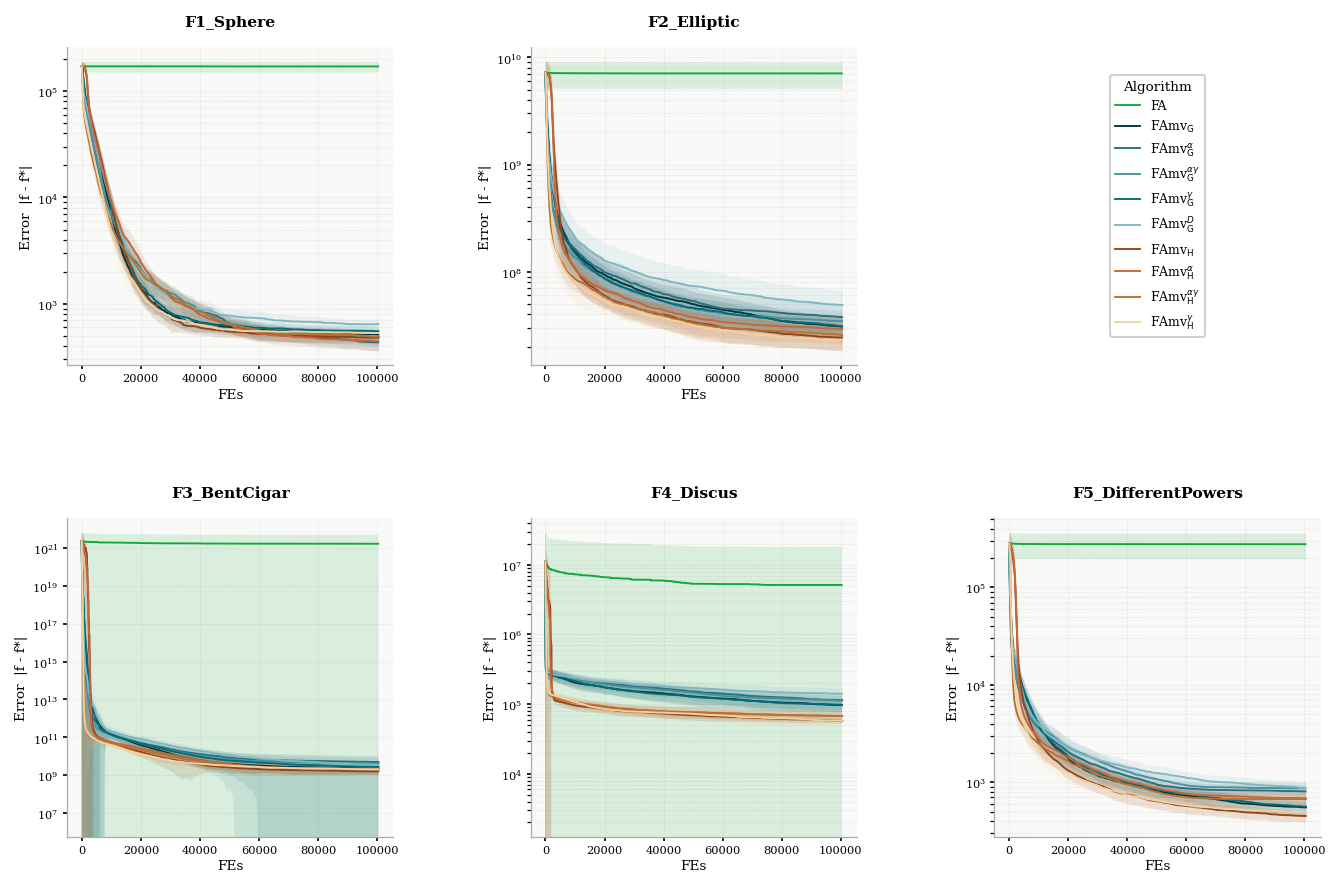}
    \caption{
    Convergence curves of the FA-based variants on \textbf{unimodal functions} in the CEC benchmark.}
    \label{fig:ab_conv1}
\end{figure}

\begin{figure}
    \centering
    \includegraphics[width=1\linewidth]{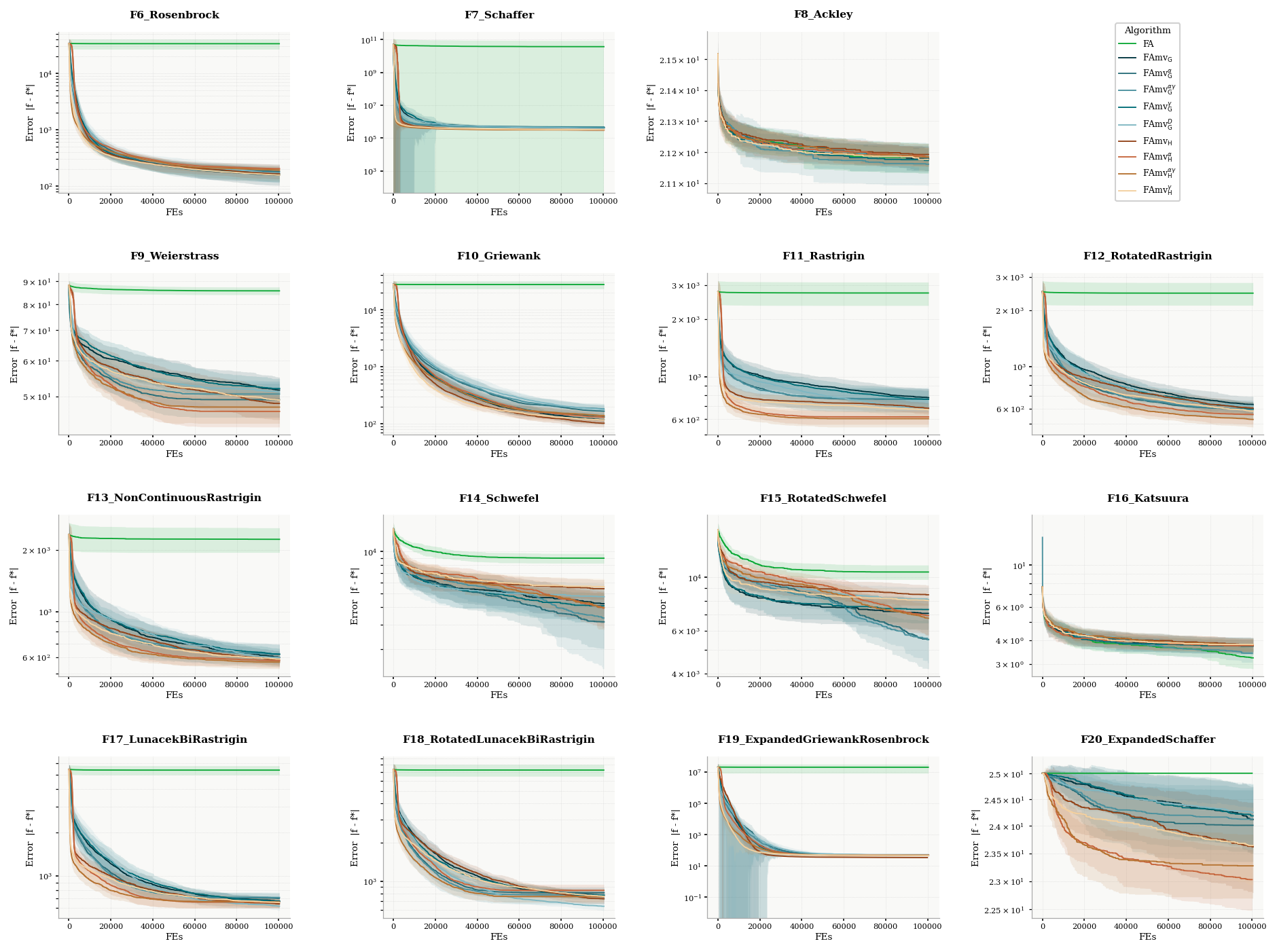}
    \caption{Convergence curves of the FA-based variants on \textbf{multimodal functions} in the CEC benchmark.}
    \label{fig:ab_conv2}
\end{figure}

\begin{figure}
    \centering
    \includegraphics[width=1\linewidth]{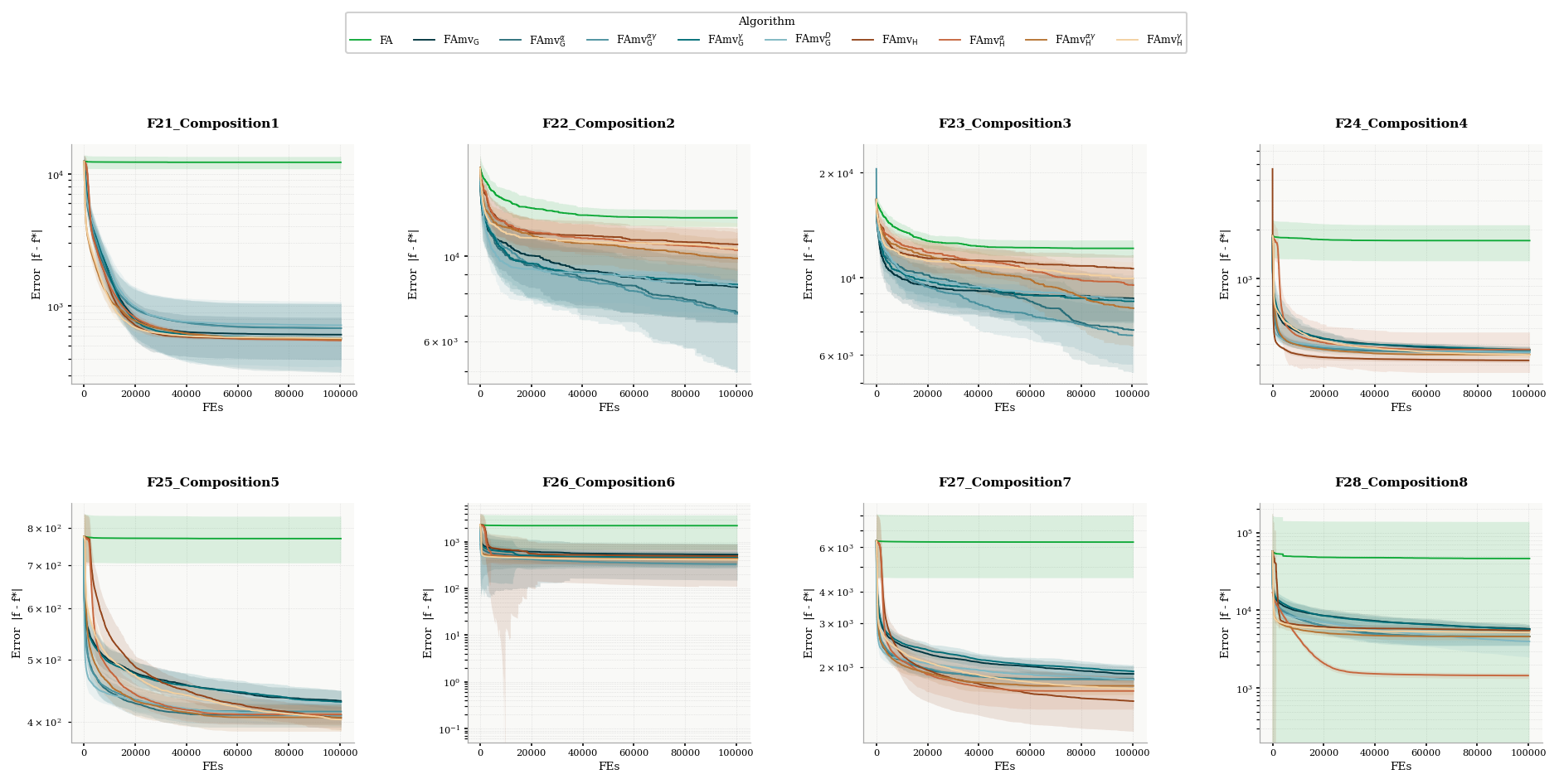}
    \caption{Convergence curves of the FA-based variants on \textbf{composition functions} in the CEC benchmark.}
    \label{fig:ab_conv3}
\end{figure}

This section presents the results obtained by the FA-based variants that investigate the proposed mixed-variable movement mechanism and the parameter adaptation strategy, as detailed in Sections~\ref{subsec:famv_movement} and~\ref{subsec:famv_adaptation}, respectively. In total, eight FA-based variants were tested to investigate the impact of certain design choices. Four FA-variants based on mixed Euclidean-Hamming (FAmv$_\mathrm{H}$, FAmv$^{\alpha}_\mathrm{H}$, FAmv$^{\alpha\gamma}_\mathrm{H}$ and FAmv$^{\gamma}_\mathrm{H}$) and four variants based on the Gower distance (FAmv$_\mathrm{G}$, FAmv$^{\alpha}_\mathrm{G}$, FAmv$^{\alpha\gamma}_\mathrm{G}$ and FAmv$^{\gamma}_\mathrm{G}$). In these definitions, the superscript indicates the parameter or parameters that are adapted. For example, ${\alpha\gamma}$ indicates both parameters are adapted simultaneously, whereas the superscript, ${\alpha}$, indicates that only ${\alpha}$ is adapted and therefore, ${\gamma}$ remains fixed. Additionally, the original FA algorithm is considered to investigate whether these FA variants for MVO problems are needed. The results on the CEC benchmark functions are summarized in Table~\ref{tab:Ablation_Results_CEC2013}. Additionally, figures of performance and convergence plots with specific results for separate problem categories are presented: unimodal (Figures~\ref{fig:ab_bp1} and~\ref{fig:ab_conv1}), multimodal (Figures~\ref{fig:ab_bp2} and~\ref{fig:ab_conv2}), and composition (Figures~\ref{fig:ab_bp3} and~\ref{fig:ab_conv3}).

From the results reported in Table~\ref{tab:Ablation_Results_CEC2013}, it can be observed that all the proposed mixed-variable variants of the Firefly Algorithm consistently outperform the classical version of the algorithm on the CEC2013 mixed-variable benchmark. In fact, the proposed variants achieve strictly better results on almost all functions. The only exception is function F8, where all algorithms, including the classical Firefly version, are statistically similar.

A more detailed analysis of the results shows that, for the unimodal functions (F1–F5), variants based on the Hamming distance are often the best-performing approaches. In particular, the Hamming-based variants (FAmv$_\mathrm{H}$) achieve the best results on functions F2, F3, F4, and F5. On function F1, the variants (FAmv$_\mathrm{H}$) are statistically similar to the best-performing algorithm.

For the multimodal functions (F6–F20), the (FAmv$_\mathrm{H}$) also appear frequently among the best-performing methods. In this group of functions, the variant using the Hamming distance with adaptive control parameters $\alpha$ and $\gamma$, FAmv$^{\alpha, \gamma}_\mathrm{H}$, tends to appear more often among the best or among results that are statistically similar. In particular, this variant achieves the best results on functions such as F11, F12, and F13, and remains statistically competitive on many other functions within this category. This behavior suggests that the adaptive parameter mechanism provides additional flexibility when dealing with the more complex landscapes typically associated with multimodal problems.

Finally, for the composition functions (F21–F28), the Hamming-based variants continue to show strong performance overall. However, the variants based on the Gower distance appear more frequently among the best-performing approaches in this group of functions, particularly when adaptive control parameters FAmv$^{\alpha}_\mathrm{G}$ and FAmv$^{\alpha, \gamma}_\mathrm{G}$ are employed. This observation suggests that these variants may offer competitive alternatives in more complex, highly composite optimization landscapes.

\begin{table*}[ht!]
\centering
\small
\renewcommand{\arraystretch}{0.85}
\setlength{\tabcolsep}{0.0cm}
\caption{ Performance of the FA-based variants evaluated on the 28 CEC benchmark functions. Performance is evaluated in terms of the absolute error (AE). The best value (lowest average) is indicated with (*), whereas statistically similar results are highlighted in \textbf{bold}.}
\label{tab:Ablation_Results_CEC2013}

\begin{tabular*}{\textwidth}{@{\extracolsep{\fill}} l l c c c c c c c c c c @{}}
\toprule
\textbf{Func.} & \textbf{Statistic} & \textbf{FA} & \textbf{FAmv$_\mathrm{G}$} & \textbf{FAmv$^{\alpha}_\mathrm{G}$} & \textbf{FAmv$^{\alpha\gamma} _\mathrm{G}$} & \textbf{FAmv$^{\gamma}_\mathrm{G}$} & \textbf{FAmv$^{D}_\mathrm{G}$} & \textbf{FAmv$_\mathrm{H}$} & \textbf{FAmv$^{\alpha}_\mathrm{H}$} & \textbf{FAmv$^{\alpha\gamma}_\mathrm{H}$} & \textbf{FAmv$^{\gamma}_\mathrm{H}$} \\
\midrule

\multirow{2}{*}{F1} & Mean & 1.71e+05 & 5.16e+02 & \textbf{4.35e+02}* & \textbf{4.79e+02} & 5.53e+02 & 6.42e+02 & \textbf{4.81e+02} & \textbf{4.47e+02} & \textbf{4.90e+02} & 5.31e+02 \\
                     & STD  & 1.92e+04 & 5.69e+01 & \textbf{7.08e+01}* & \textbf{7.92e+01} & 6.57e+01 & 7.28e+01 & \textbf{5.60e+01} & \textbf{8.43e+01} & \textbf{7.80e+01} & 6.28e+01 \\

\multirow{2}{*}{F2} & Mean & 7.10e+09 & 3.08e+07 & 3.77e+07 & 3.46e+07 & 3.09e+07 & 4.89e+07 & \textbf{2.43e+07} & \textbf{2.92e+07} & \textbf{2.55e+07} & \textbf{2.30e+07}* \\
                     & STD  & 1.98e+09 & 8.63e+06 & 1.25e+07 & 8.99e+06 & 6.51e+06 & 1.86e+07 & \textbf{5.69e+06} & \textbf{8.08e+06} & \textbf{7.25e+06} & \textbf{4.64e+06}* \\

\multirow{2}{*}{F3} & Mean & 1.60e+21 & \textbf{2.25e+09} & 5.10e+09 & 4.14e+09 & 2.84e+09 & 3.68e+09 & \textbf{1.63e+09}* & \textbf{2.07e+09} & \textbf{2.09e+09} & \textbf{2.12e+09} \\
                     & STD  & 3.08e+21 & \textbf{1.23e+09} & 5.22e+09 & 4.55e+09 & 1.07e+09 & 1.66e+09 & \textbf{5.47e+08}* & \textbf{8.97e+08} & \textbf{9.82e+08} & \textbf{1.00e+09} \\

\multirow{2}{*}{F4} & Mean & 5.15e+06 & 9.68e+04 & 1.14e+05 & 1.12e+05 & 9.78e+04 & 1.41e+05 & \textbf{5.75e+04}* & \textbf{6.79e+04} & \textbf{6.75e+04} & \textbf{5.78e+04} \\
                     & STD  & 1.30e+07 & 1.95e+04 & 1.76e+04 & 2.99e+04 & 2.14e+04 & 2.84e+04 & \textbf{5.46e+03}* & \textbf{1.26e+04} & \textbf{1.21e+04} & \textbf{9.94e+03} \\

\multirow{2}{*}{F5} & Mean & 2.78e+05 & \textbf{5.52e+02} & 8.05e+02 & 8.64e+02 & \textbf{5.62e+02} & 8.63e+02 & \textbf{4.50e+02}* & 6.87e+02 & 6.70e+02 & \textbf{4.91e+02} \\
                     & STD  & 8.13e+04 & \textbf{8.21e+01} & 1.19e+02 & 1.41e+02 & \textbf{1.03e+02} & 1.17e+02 & \textbf{5.97e+01}* & 1.15e+02 & 1.15e+02 & \textbf{7.77e+01} \\

\midrule
\multicolumn{9}{l}{\textbf{Basic multimodal functions (F6--F20)}} \\
\midrule

\multirow{2}{*}{F6} & Mean & 3.34e+04 & \textbf{1.57e+02} & \textbf{1.81e+02} & \textbf{1.71e+02} & \textbf{1.65e+02} & \textbf{1.68e+02} & \textbf{1.65e+02} & 2.04e+02 & \textbf{1.92e+02} & \textbf{1.57e+02}* \\
                     & STD  & 6.74e+03 & \textbf{5.45e+01} & \textbf{6.22e+01} & \textbf{5.43e+01} & \textbf{4.24e+01} & \textbf{5.37e+01} & \textbf{3.11e+01} & 3.96e+01 & \textbf{3.57e+01} & \textbf{3.49e+01}* \\

\multirow{2}{*}{F7} & Mean & 3.68e+10 & 4.27e+05 & 4.56e+05 & 4.60e+05 & 4.21e+05 & 3.98e+05 & \textbf{3.02e+05}* & \textbf{3.27e+05} & \textbf{3.16e+05} & \textbf{3.17e+05} \\
                     & STD  & 4.54e+10 & 5.82e+04 & 6.79e+04 & 7.19e+04 & 5.68e+04 & 8.36e+04 & \textbf{3.18e+04}* & \textbf{4.89e+04} & \textbf{4.22e+04} & \textbf{3.22e+04} \\

\multirow{2}{*}{F8} & Mean & \textbf{2.12e+01} & \textbf{2.12e+01} & \textbf{2.12e+01} & \textbf{2.12e+01}* & \textbf{2.12e+01} & \textbf{2.12e+01} & \textbf{2.12e+01} & \textbf{2.12e+01} & \textbf{2.12e+01} & \textbf{2.12e+01} \\
                     & STD  & \textbf{5.00e-02} & \textbf{4.00e-02} & \textbf{4.00e-02} & \textbf{7.00e-02}* & \textbf{4.00e-02} & \textbf{3.00e-02} & \textbf{4.00e-02} & \textbf{3.00e-02} & \textbf{3.00e-02} & \textbf{4.00e-02} \\

\multirow{2}{*}{F9} & Mean & 8.56e+01 & 5.16e+01 & \textbf{4.92e+01} & 5.06e+01 & 5.20e+01 & 5.11e+01 & \textbf{4.82e+01} & \textbf{4.63e+01}* & \textbf{4.73e+01} & \textbf{4.91e+01} \\
                     & STD  & 1.67e+00 & 3.50e+00 & \textbf{4.23e+00} & 3.41e+00 & 2.68e+00 & 2.83e+00 & \textbf{3.84e+00} & \textbf{3.54e+00}* & \textbf{3.77e+00} & \textbf{3.06e+00} \\

\multirow{2}{*}{F10} & Mean & 2.75e+04 & \textbf{1.18e+02} & 1.67e+02 & 1.83e+02 & 1.32e+02 & 1.78e+02 & \textbf{1.03e+02}* & 1.31e+02 & 1.40e+02 & \textbf{1.18e+02} \\
                     & STD  & 4.31e+03 & \textbf{2.12e+01} & 4.91e+01 & 4.34e+01 & 2.25e+01 & 3.28e+01 & \textbf{1.48e+01}* & 3.08e+01 & 2.97e+01 & \textbf{2.07e+01} \\

\multirow{2}{*}{F11} & Mean & 2.74e+03 & 7.77e+02 & 7.63e+02 & 7.65e+02 & 7.62e+02 & 6.83e+02 & 6.84e+02 & \textbf{6.17e+02} & \textbf{6.04e+02}* & \textbf{6.71e+02} \\
                     & STD  & 3.83e+02 & 8.75e+01 & 1.03e+02 & 1.14e+02 & 8.51e+01 & 6.66e+01 & 4.72e+01 & \textbf{5.30e+01} & \textbf{5.77e+01}* & \textbf{4.36e+01} \\

\multirow{2}{*}{F12} & Mean & 2.46e+03 & 6.32e+02 & 5.94e+02 & 5.94e+02 & 6.00e+02 & 6.20e+02 & 6.04e+02 & \textbf{5.59e+02} & \textbf{5.26e+02}* & \textbf{5.82e+02} \\
                     & STD  & 3.38e+02 & 6.59e+01 & 6.01e+01 & 7.16e+01 & 4.76e+01 & 3.62e+01 & 4.78e+01 & \textbf{6.24e+01} & \textbf{4.68e+01}* & \textbf{5.58e+01} \\

\multirow{2}{*}{F13} & Mean & 2.25e+03 & \textbf{5.99e+02} & \textbf{6.21e+02} & \textbf{6.19e+02} & 6.22e+02 & \textbf{5.98e+02} & \textbf{5.80e+02} & \textbf{5.79e+02} & \textbf{5.69e+02}* & \textbf{5.88e+02} \\
                     & STD  & 3.08e+02 & \textbf{4.22e+01} & \textbf{7.09e+01} & \textbf{7.71e+01} & 4.14e+01 & \textbf{3.60e+01} & \textbf{3.43e+01} & \textbf{4.27e+01} & \textbf{4.13e+01}* & \textbf{4.46e+01} \\

\multirow{2}{*}{F14} & Mean & 8.95e+03 & \textbf{4.23e+03} & \textbf{3.13e+03}* & \textbf{3.34e+03} & \textbf{4.08e+03} & 4.74e+03 & 5.41e+03 & \textbf{3.93e+03} & \textbf{3.97e+03} & 5.58e+03 \\
                     & STD  & 7.19e+02 & \textbf{1.17e+03} & \textbf{1.11e+03}* & \textbf{1.89e+03} & \textbf{9.70e+02} & 1.16e+03 & 8.71e+02 & \textbf{1.08e+03} & \textbf{1.20e+03} & 7.81e+02 \\

\multirow{2}{*}{F15} & Mean & 1.05e+04 & 7.09e+03 & \textbf{5.53e+03} & \textbf{5.52e+03}* & 7.35e+03 & 8.08e+03 & 8.46e+03 & 6.94e+03 & \textbf{6.77e+03} & 8.01e+03 \\
                     & STD  & 7.18e+02 & 1.01e+03 & \textbf{1.34e+03} & \textbf{1.10e+03}* & 8.84e+02 & 9.90e+02 & 8.05e+02 & 1.16e+03 & \textbf{1.01e+03} & 7.64e+02 \\

\multirow{2}{*}{F16} & Mean & \textbf{3.24e+00} & 3.75e+00 & 3.80e+00 & \textbf{3.42e+00}* & 3.72e+00 & 3.78e+00 & 3.79e+00 & 3.74e+00 & 3.81e+00 & 3.81e+00 \\
                     & STD  & \textbf{4.00e-01} & 3.80e-01 & 3.10e-01 & \textbf{3.60e-01}* & 3.20e-01 & 3.60e-01 & 2.90e-01 & 3.00e-01 & 3.20e-01 & 3.00e-01 \\

\multirow{2}{*}{F17} & Mean & 5.39e+03 & 6.69e+02 & 7.00e+02 & 7.10e+02 & 6.69e+02 & \textbf{6.18e+02}* & \textbf{6.42e+02} & \textbf{6.58e+02} & \textbf{6.62e+02} & \textbf{6.57e+02} \\
                     & STD  & 3.88e+02 & 3.33e+01 & 6.06e+01 & 5.91e+01 & 3.81e+01 & \textbf{4.01e+01}* & \textbf{3.88e+01} & \textbf{5.87e+01} & \textbf{5.54e+01} & \textbf{4.26e+01} \\

\multirow{2}{*}{F18} & Mean & 7.29e+03 & 7.75e+02 & 8.12e+02 & 7.82e+02 & 7.73e+02 & \textbf{6.38e+02}* & 7.29e+02 & 8.47e+02 & 7.58e+02 & 7.62e+02 \\
                     & STD  & 7.63e+02 & 5.71e+01 & 6.78e+01 & 1.18e+02 & 5.84e+01 & \textbf{4.57e+01}* & 6.18e+01 & 9.59e+01 & 7.27e+01 & 5.54e+01 \\

\multirow{2}{*}{F19} & Mean & 2.08e+07 & 4.91e+01 & 5.25e+01 & 5.05e+01 & 4.98e+01 & 4.81e+01 & \textbf{3.48e+01}* & 4.96e+01 & 4.86e+01 & 4.76e+01 \\
                     & STD  & 1.22e+07 & 2.73e+00 & 5.46e+00 & 4.46e+00 & 3.07e+00 & 2.74e+00 & \textbf{1.49e+00}* & 3.59e+00 & 2.73e+00 & 2.77e+00 \\

\multirow{2}{*}{F20} & Mean & 2.50e+01 & 2.41e+01 & 2.40e+01 & 2.41e+01 & 2.42e+01 & 2.42e+01 & \textbf{2.36e+01} & \textbf{2.30e+01}* & \textbf{2.33e+01} & \textbf{2.36e+01} \\
                     & STD  & 0.00e+00 & 6.20e-01 & 6.70e-01 & 5.90e-01 & 6.00e-01 & 5.40e-01 & \textbf{8.20e-01} & \textbf{5.40e-01}* & \textbf{5.80e-01} & \textbf{7.10e-01} \\

\midrule
\multicolumn{9}{l}{\textbf{Composition functions (F21--F28)}} \\
\midrule

\multirow{2}{*}{F21} & Mean & 1.22e+04 & 6.07e+02 & \textbf{6.79e+02} & \textbf{6.79e+02} & 5.72e+02 & 7.15e+02 & \textbf{5.60e+02} & \textbf{5.49e+02}* & \textbf{5.59e+02} & 5.70e+02 \\
                     & STD  & 1.30e+03 & 2.14e+02 & \textbf{3.65e+02} & \textbf{3.62e+02} & 9.50e+00 & 3.65e+02 & \textbf{8.36e+00} & \textbf{9.66e+00}* & \textbf{1.27e+01} & 9.67e+00 \\

\multirow{2}{*}{F22} & Mean & 1.26e+04 & \textbf{8.31e+03} & \textbf{7.15e+03} & \textbf{7.08e+03}* & \textbf{8.45e+03} & \textbf{8.39e+03} & 1.08e+04 & 1.04e+04 & 9.89e+03 & 1.05e+04 \\
                     & STD  & 6.44e+02 & \textbf{1.62e+03} & \textbf{2.15e+03} & \textbf{2.11e+03}* & \textbf{1.72e+03} & \textbf{1.33e+03} & 1.10e+03 & 1.20e+03 & 1.29e+03 & 1.78e+03 \\

\multirow{2}{*}{F23} & Mean & 1.21e+04 & 8.71e+03 & \textbf{7.07e+03} & \textbf{6.82e+03}* & 8.55e+03 & 8.62e+03 & 1.06e+04 & 9.50e+03 & \textbf{8.16e+03} & 9.93e+03 \\
                     & STD  & 6.86e+02 & 1.23e+03 & \textbf{1.73e+03} & \textbf{1.18e+03}* & 1.14e+03 & 1.54e+03 & 1.02e+03 & 9.89e+02 & \textbf{1.80e+03} & 1.16e+03 \\

\multirow{2}{*}{F24} & Mean & 1.71e+03 & 3.69e+02 & 3.60e+02 & 3.63e+02 & 3.70e+02 & 3.57e+02 & \textbf{3.19e+02}* & 3.72e+02 & 3.46e+02 & 3.45e+02 \\
                     & STD  & 4.22e+02 & 1.87e+01 & 1.53e+01 & 1.61e+01 & 1.64e+01 & 1.43e+01 & \textbf{9.44e+00}* & 1.02e+02 & 1.44e+01 & 1.32e+01 \\

\multirow{2}{*}{F25} & Mean & 7.70e+02 & 4.31e+02 & \textbf{4.10e+02} & \textbf{4.15e+02} & 4.29e+02 & \textbf{4.08e+02} & \textbf{4.06e+02} & \textbf{4.11e+02} & \textbf{4.06e+02} & \textbf{4.03e+02}* \\
                     & STD  & 6.41e+01 & 1.57e+01 & \textbf{1.76e+01} & \textbf{1.87e+01} & 1.85e+01 & \textbf{1.31e+01} & \textbf{1.69e+01} & \textbf{1.43e+01} & \textbf{1.98e+01} & \textbf{1.58e+01}* \\

\multirow{2}{*}{F26} & Mean & 2.20e+03 & 5.28e+02 & 4.59e+02 & \textbf{3.29e+02}* & 4.51e+02 & 4.30e+02 & \textbf{4.90e+02} & \textbf{4.30e+02} & \textbf{4.30e+02} & \textbf{4.09e+02} \\
                     & STD  & 1.53e+03 & 3.79e+02 & 1.52e+02 & \textbf{6.26e+01}* & 1.69e+02 & 9.71e+00 & \textbf{3.82e+02} & \textbf{1.06e+02} & \textbf{9.19e+01} & \textbf{4.01e+01} \\

\multirow{2}{*}{F27} & Mean & 6.28e+03 & 1.88e+03 & 1.79e+03 & 1.79e+03 & 1.92e+03 & 1.79e+03 & \textbf{1.47e+03}* & \textbf{1.61e+03} & \textbf{1.69e+03} & \textbf{1.64e+03} \\
                     & STD  & 1.75e+03 & 1.41e+02 & 1.19e+02 & 1.37e+02 & 1.22e+02 & 1.32e+02 & \textbf{3.53e+02}* & \textbf{2.47e+02} & \textbf{1.32e+02} & \textbf{1.32e+02} \\

\multirow{2}{*}{F28} & Mean & 4.64e+04 & 5.77e+03 & 4.63e+03 & 4.61e+03 & 5.71e+03 & 3.97e+03 & 5.51e+03 & \textbf{1.45e+03}* & 4.60e+03 & 5.19e+03 \\
                     & STD  & 9.05e+04 & 7.57e+02 & 1.12e+03 & 1.12e+03 & 6.24e+02 & 1.50e+03 & 5.21e+02 & \textbf{1.19e+02}* & 4.91e+02 & 3.30e+02 \\

\midrule
\textbf{Count } &
& \textbf{2}  \textbf{(0)}
& \textbf{8} \textbf{(0)}
& \textbf{11} \textbf{(2)}
& \textbf{12} \textbf{(6)}
& \textbf{5}  \textbf{(0)}
& \textbf{7}  \textbf{(2)}
& \textbf{19} \textbf{(8)}
& \textbf{18} \textbf{(4)}
& \textbf{20} \textbf{(3)}
& \textbf{17} \textbf{(3)} \\
\bottomrule
\end{tabular*}
\end{table*}

\end{document}